\documentclass[12pt,english]{article}

\usepackage{amsmath,amsfonts,amsthm,amssymb}
\usepackage{array}
\usepackage{setspace, multirow}
\usepackage{Tabbing}
\usepackage{graphicx,float}
\usepackage{slashbox}
\usepackage{booktabs}
\usepackage{algorithmic}
\usepackage[authoryear]{natbib}
\usepackage[colorlinks,citecolor=blue,
    linkcolor=black]{hyperref}
\hypersetup{bookmarksnumbered}
\usepackage{geometry}
\allowdisplaybreaks[3]

\topmargin by -\headheight \advance
\topmargin by -\headsep

\evensidemargin
\oddsidemargin
\newcommand{\ra}[1]{\renewcommand{\arraystretch}{#1}}
\DeclareMathOperator*{\argmin}{argmin}

\graphicspath{ {./bw_images/} }

\usepackage[width=\textwidth ,font={small},labelfont={color=blue,bf,up},
labelsep=endash, format=plain, labelsep=period,
textfont=it]{caption}

\newtheorem{Lemma}{Lemma}
\newtheorem{theorem}{Theorem}
\numberwithin{equation}{section}

\floatstyle{ruled}
\newfloat{algorithm}{tbp}{loa}
\providecommand{\algorithmname}{Algorithm}
\floatname{algorithm}{\protect\algorithmname}

\newcommand{\bs}{\boldsymbol }

\begin{document}
\title{\LARGE \bf Another Look at DWD: Thrifty Algorithm and Bayes Risk Consistency in RKHS}

\author{\sc{Boxiang Wang\thanks{School of Statistics, University of Minnesota.}} and \sc{Hui Zou} \thanks{Corresponding author, zouxx019@umn.edu. School of Statistics, University of Minnesota.}\\
}
\date{August 21, 2015}

\maketitle
\begin{abstract}
Distance weighted discrimination (DWD) is a margin-based classifier with an interesting geometric motivation. DWD was originally proposed as a superior alternative to the support vector machine (SVM), however DWD is yet to be popular compared with the SVM. The main reasons are twofold. First, the state-of-the-art algorithm for solving DWD is based on the second-order-cone programming (SOCP), while the SVM is a quadratic programming problem which is much more efficient to solve. Second, the current statistical theory of DWD mainly focuses on the linear DWD for the high-dimension-low-sample-size setting and data-piling, while the learning theory for the SVM mainly focuses on the Bayes risk consistency of the kernel SVM. In fact, the Bayes risk consistency of DWD is presented as an open problem in the original DWD paper. In this work, we advance the current understanding of DWD from both computational and theoretical perspectives. We propose a novel efficient algorithm for solving DWD, and our algorithm can be several hundred times faster than the existing state-of-the-art algorithm based on the SOCP. In addition, our algorithm can handle the generalized DWD, while the SOCP algorithm only works well for a special DWD but not the generalized DWD. Furthermore, we consider a natural kernel DWD in a reproducing kernel Hilbert space and then establish the Bayes risk consistency of the kernel DWD. We compare DWD and the SVM on several benchmark data sets and show that the two have comparable classification accuracy, but DWD equipped with our new algorithm can be much faster to compute than the SVM.

\end{abstract}

\section{Introduction}
Binary classification problems appear from diverse practical applications, such as, financial fraud detection, spam email classification, medical diagnosis with genomics data, drug response modeling, among many others. In these classification problems, the goal is to predict class labels based on a given set of variables. Suppose that we observe a training data set consisting of $n$ pairs, where $\{(\bs{x}_i, y_i)\}_{i=1}^n$, $\bs{x}_i\in\mathbb{R}^p$, and $y_i \in \{-1, 1\}$. A classifier fits a discriminant function $f$ and constructs a classification rule to classify data point $\bs{x_i}$ to either class $1$ or class $-1$ according to the sign of $f(\bs{x}_i)$. The decision boundary is given by $\{\bs{x}:f(\bs{x})=0\}$. Two canonical classifiers are linear discriminant analysis and logistic regression. Modern classification algorithms can produce flexible non-linear decision boundaries with high accuracy. The two most popular approaches are ensemble learning and support vector machines/kernel machines. Ensemble learning such as boosting \citep{FreundSchapire1997} and random forest \citep{Breiman2001} combine many weak learners like decision trees into a powerful one. The support vector machine (SVM) \citep{Vapnik1995, Vapnik1998} fits an optimal separating hyperplane in the extended kernel feature space which is non-linear in the original covariate spaces. In a recent extensive numerical study by \cite{FernandezEtAl2014}, the kernel SVM is shown to be one of the best among 179 commonly used classifiers.

Motivated by ``data-piling" in the high-dimension-low-sample-size problems, \cite{MarronEtAl2007} invented a new classification algorithm named distance weighted discrimination (DWD) that retains the elegant geometric interpretation of the SVM and delivers competitive performance. Since then much work has been devoted to the development of DWD. The readers are referred to \cite{Marron2015} for an up-to-date list of work on DWD.
On the other hand, we notice that DWD has not attained the popularity it deserves. We can think of two reasons for that. First,
the current state-of-the-art algorithm for DWD is based on second-order-cone programming (SOCP) proposed in \cite{MarronEtAl2007}. SOCP was an essential part of the DWD development. As acknowledged in \cite{MarronEtAl2007}, SOCP was then much less well-known than quadratic programming, even in optimization. Furthermore, SOCP is generally more computationally demanding than quadratic programming. There are two existing implementations of the SOCP algorithm: \cite{Marron2013} in Matlab and \cite{HuangEtAl2012} in R. With these two implementations, we find that DWD is usually more time-consuming than the SVM.
Therefore, SOCP contributes to both the success and unpopularity of DWD.
Second, the kernel extension of DWD and the corresponding kernel learning theory are under-developed compared to the kernel SVM. Although \cite{MarronEtAl2007} proposed a version of non-linear DWD by mimicking the kernel trick used for deriving the kernel SVM, theoretical justification of such a kernel DWD is still absent. On the contrary, the kernel SVM as well as the kernel logistic regression \citep{WahbaEtAl1994, ZhuHastie2005} have mature theoretical understandings built upon the theory of reproducing kernel Hilbert space (RKHS) \citep{Wahba1999, HastieEtAl2009}. Most learning theories of DWD succeed to \cite{HallEtAl2005}'s geometric view of HDLSS data and assume that $p \to \infty$ and $n$ is fixed, as opposed to the learning theory for the SVM where $n \to \infty$ and $p$ is fixed. We are not against the fixed $n$ and $p \to \infty$ theory but it would be desirable to develop the canonical learning theory for the kernel DWD when $p$ is fixed and $n \to \infty$. In fact, how to establish the Bayes risk consistency of the DWD and kernel DWD was proposed as an open research problem in the original DWD paper \citep{MarronEtAl2007}. Nearly a decade later, the problem still remains open. 

In this paper, we aim to resolve the aforementioned issues. We show that the kernel DWD in a RKHS has the Bayes risk consistency property if a universal kernel is used. This result should convince those who are less familiar with DWD to treat the kernel DWD as a serious competitor to the kernel SVM. To popularize the DWD, it is also important to allow practitioners to easily try DWD collectively with the SVM in real applications. To this end, we develop a novel fast algorithm to solve the linear and kernel DWD by using the majorization-minimization (MM) principle. Compared with the SOCP algorithm, our new algorithm has multiple advantages. First, our algorithm is much faster than the SOCP algorithm. In some examples, our algorithm can be several hundred times faster. Second, DWD equipped with our algorithm can be faster than the SVM. Third, our algorithm is easier to understand than the SOCP algorithm, especially for those who are not familiar with semi-definite and second-order-cone programming. This could help demystify the DWD and hence may increase its popularity. 

\begin{figure}[t]
\begin{minipage}[b]{\linewidth}
\centering
\includegraphics[width=\textwidth]{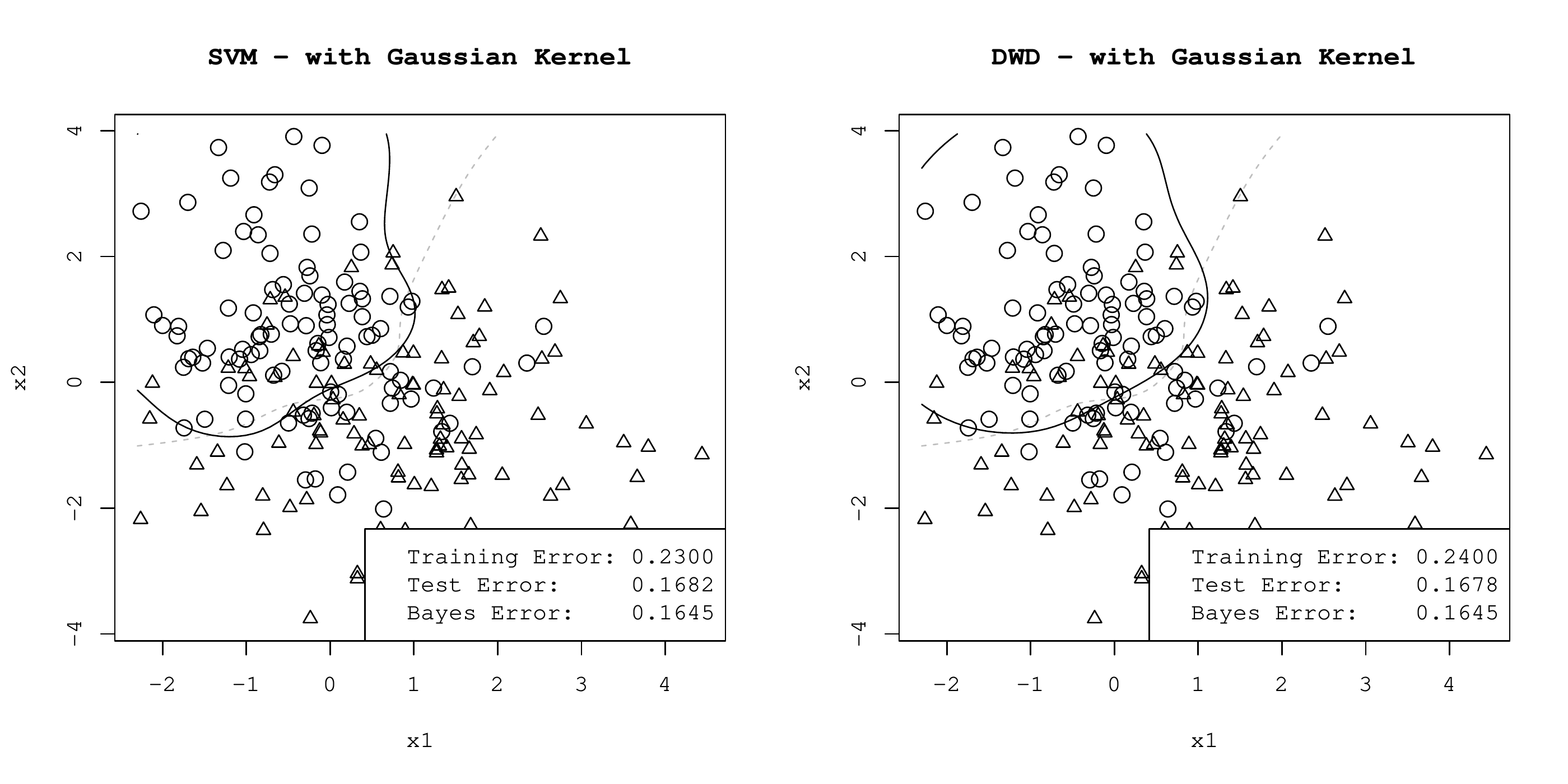}
\caption{Nonlinear SVM and DWD with Gaussian kernel. 
The broken curves are the Bayes decision boundary. The R package \texttt{kerndwd} used 2.396 second to solve the kernel DWD, and \texttt{kernlab} took 7.244 second to solve the kernel SVM. The timings include tuning parameters and they are averaged over 100 runs.
}
\label{fig:svm_dwd}
\end{minipage}
\end{figure}

To give a quick demonstration, we use a simulation example to compare the kernel DWD and the kernel SVM.  We drew 10 centers $\{\bs \mu_{k+} \}$ from $N((1,0)^T, \bs I)$. For each data point in the positive class, we randomly picked up a center $\bs\mu_{k+}$ and then generated the point from $N(\bs\mu_{k+}, \bs I/5)$. The negative class was assembled in the same way except that 10 centers $\bs \mu_{k-}$ were drawn from $N((0,1)^T, \bs I)$. For this model the Bayes rule is nonlinear
\footnote{The Bayes decision boundary is a curve:
$\left\{ \bs z:
\sum_{k} \exp\left(-5||\bs z - \bs \mu_{k+}||^2/2 \right) = \sum_{k}  \exp\left(-5||\bs z - \bs \mu_{k-}||^2/2 \right)
\right\}.$}. Figure~\ref{fig:svm_dwd} displays the training data from the simulation model where 100 observations are from the positive class (plotted as triangles) and another 100 observations are from the negative class (plotted as circles). We fitted the SVM and DWD using Gaussian kernels. We have implemented our new algorithm for DWD in a publicly available R package \texttt{kerndwd}. We computed the kernel SVM by using the R package \texttt{kernlab} \citep{KaratzoglouEtAl2004}.
We recorded their training errors and test errors. 
From Figure~\ref{fig:svm_dwd}, we observe that like the kernel SVM, the kernel DWD has a test error close to the Bayes error, which is consistent with the Bayes risk consistency property of the kernel DWD established in section~\ref{sec:learningtheory0}.
Notably, the kernel DWD is about three times as fast as the kernel SVM in this example.

The rest of the paper is organized as follows. To be self-contained, we first review the SVM and DWD in section~\ref{sec:review}.
We then derive the novel algorithm for DWD in section~\ref{sec:computation}. We introduce the kernel DWD in a reproducing kernel Hilbert space and establish the learning theory of kernel DWD in section~\ref{sec:kerDWD}. Real data examples are given in section~\ref{sec:real_data} to compare DWD and the SVM. Technical proofs are provided in the appendix.

\section{Review of SVMs and DWD}
\label{sec:review}

\subsection{SVM}
\label{sec:review_SVM}
The introduction of the SVM usually begins with its geometric interpretation as a maximum margin classifier \citep{Vapnik1995}. Consider a case when two classes are separable by a hyperplane $\{\bs x: f(x) = \omega_0 + \bs x^T \bs\omega=0\}$ such that $y_i(\omega_0 + \bs{x}_i^T \bs\omega)$ are all non-negative. Without loss of generality, we assume that $\bs\omega$ is a unit vector, i.e., $\bs\omega^T \bs\omega=1$, and we observe that each $d_i\equiv y_i(\omega_0 + \bs{x}_i^T \bs\omega)$ is equivalent to the Euclidean distance between the data point $\bs x_i$ and the hyperplane. The reason is that $d_i = (\bs x_i - \bs x_0)^T \bs\omega$ and $\omega_0 + \bs x_0^T \bs \omega = 0$, where $\bs x_0$ is any data point on the hyperplane and $\bs \omega$ is the unit normal vector. The SVM classifier is defined as the optimal separating hyperplane that maximizes the smallest distance of each data point to the separating hyperplane. Mathematically, the SVM can be written as the following optimization problem (for the separable data case):
\begin{equation}
\begin{aligned}
\max_{\omega_0, \bs\omega}& \;\;\; \min{d_i},\\
\text{ subject to }& \;\;\; d_i = y_i(\omega_0 + \bs x_i^T\bs\omega) \ge 0, \ \forall i, \text{ and } \bs\omega^T \bs\omega=1.
\label{eq:sep_SVM}
\end{aligned}
\end{equation}
The smallest distance $\min d_i$ is called the \textit{margin}, and the SVM is thereby regarded as a \textit{large-margin classifier}. The data points closest to the hyperplane, i.e., $d_i = \min d_i$, are dubbed the \textit{support vectors}.

In general, the two classes are not separable, and thus $y_i(\omega_0 + \bs x_i^T \bs \omega)$ cannot be non-negative for all $i=1,\ldots, n$. To handle this issue, non-negative slack variables $\eta_i, \ 1 \le i \le n$, are introduced to ensure all $y_i(\omega_0 + \bs x_i^T \bs \omega) + \eta_i$ to be non-negative. With these slack variables, the optimization problem \eqref{eq:sep_SVM} is generalized as follows,
\begin{equation}
\begin{aligned}
\max_{\omega_0, \bs\omega}&\;\;\; \min{d_i},\\
\text{ subject to }&\;\;\; d_i = y_i(\omega_0 + \bs x_i^T\bs\omega) + \eta_i \ge 0, \ \forall i,\\
&\;\;\; \eta_i \ge 0,\ \forall i, \ \sum_{i=1}^n \eta_i < \text{constant}, \text{ and } \bs\omega^T \bs\omega=1.
\label{eq:nonsep_SVM}
\end{aligned}
\end{equation}

To compute SVMs, the optimization problem \eqref{eq:nonsep_SVM} is usually rephrased as an equivalent quadratic programming (QP) problem,
\begin{equation}
\begin{aligned}
\min_{\beta_0, \bs\beta}&\;\;\; \left[\dfrac{1}{2}\bs\beta^T \bs\beta + c\sum_{i=1}^n \xi_i \right],\\
\text{ subject to }&\;\;\; y_i(\beta_0 + \bs x_i^T\bs\beta) + \xi_i \ge 1,\ \xi_i \ge 0, \ \forall i,
\label{eq:quad_SVM}
\end{aligned}
\end{equation}
and it can be solved by maximizing its Lagrange dual function,
\begin{equation}
\begin{aligned}
\max_{\mu_i}&\;\;\; \left[\sum_{i=1}^n \mu_i - \dfrac{1}{2}\sum_{i=1}^n \sum_{i'=1}^n \mu_i \mu_{i'}y_i y_{i'} \langle \bs x_i,  \bs x_{i'}\rangle\right], \\
\text{ subject to }&\;\;\; \mu_i \ge 0 \text{ and }\sum_{i=1}^n \mu_i y_i = 0.
\label{eq:dual}
\end{aligned}
\end{equation}
By solving \eqref{eq:dual}, one can show that the solution of \eqref{eq:quad_SVM} has the form
\begin{equation}
\begin{aligned}
\hat{\bs\beta} = \sum_{i=1}^n \hat{\mu}_i y_i \bs x_i, \text{ and thus }
\hat{f}(\bs x) = \hat{\beta}_0 + \sum_{i=1}^n \hat{\mu}_i y_i \langle \bs x, \bs x_i \rangle,
\label{eq:soln_SVM}
\end{aligned}
\end{equation}
$\hat{\mu}_i$ being zero only when $\bs x_i$ lies on the support vectors.

One widely used method to extend the linear SVM to non-linear classifiers is the kernel method \citep{AizermanEtAl1964}, which replaces the dot product $\langle \bs x_i, \bs x_{i'} \rangle$ in the Lagrange dual problem \eqref{eq:dual} with a kernel function $K(\bs x_i, \bs x_i')$, and hence the solution has the form
$$
\hat{f}(\bs x) = \hat{\beta}_0 + \bs x^T \hat{\bs\beta} = \hat{\beta}_0 + \sum_{i=1}^n \hat{\mu}_i y_i K(\bs x, \bs x_i).
$$
Some popular examples of the kernel function $K$ include:  $K(\bs x, \bs x') = \langle \bs x, \bs x' \rangle$ (linear kernel), 
$K(\bs x, \bs x') = \left(a + \langle \bs x, \bs x' \rangle \right)^d$ (polynomial kernel), and $K(\bs x, \bs x') = \exp(-\sigma||\bs x - \bs x'||_2^2)$ (Gaussian kernel), among others.
%

\subsection{DWD}
\subsubsection{Motivation}
\begin{figure}[t]
\begin{minipage}[b]{\linewidth}
\centering
\includegraphics[width=\textwidth]{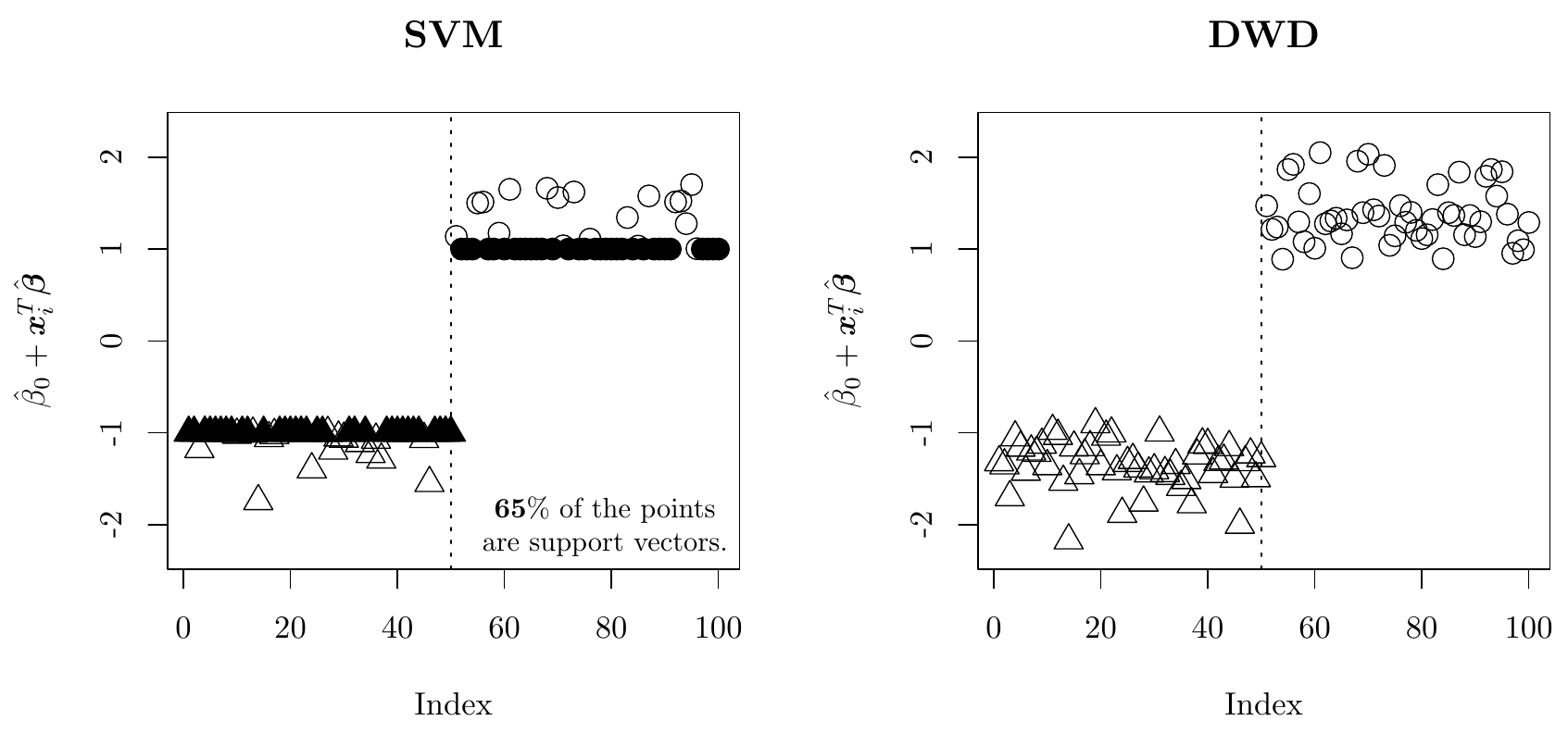}
\caption{A toy example illustrating the data-piling. Values $\hat{\beta}_0 + \bs x_i^T \hat{\bs\beta}$ are plotted for SVM and DWD. Indices 1 to 50 represent negative class (triangles) and indices 51 to 100 are for positive class (circles). In the left panel, data points belonging to the support vectors are depicted as solid circles and triangles.}
\label{fig:dp}
\end{minipage}
\end{figure}

Distance weighted discrimination was originally proposed by \cite{MarronEtAl2007} to resolve the \textit{data-piling} issue.
 \cite{MarronEtAl2007} observed that many data points become support vectors when the SVM is applied on the so-called high-dimension-low-sample-size (HDLSS) data, and \cite{MarronEtAl2007} coined the term {data-piling} to describe this phenomenon. We delineate it in Figure~\ref{fig:dp} through a simulation example. Let $\bs\mu = (3, 0, \ldots, 0)$ be a $200$-dimension vector. We generated $50$ points (indexed from $1$ to $50$ and represented as triangles) from $N(-\bs\mu, \bs I_p)$ as the negative class and another $50$ points (indexed from $51$ to $100$ and represented as circles) from $N(\bs\mu, \bs I_p)$ as the positive class. We computed $\hat{\beta}_0$ and $\hat{\bs\beta}$ for SVM \eqref{eq:quad_SVM}. In the left panel of Figure~\ref{fig:dp}, we plotted $\hat{\beta}_0 + \bs x_i^T \hat{\bs\beta}$ for each data point, and we portrayed the support vectors by solid triangles and circles. We observe that $65$ out of $100$ data points become support vectors. The right panel of Figure~\ref{fig:dp} corresponds to DWD (will be defined shortly), where data-piling is attenuated. A real example revealing the data-piling can be seen in Figure 1 of \cite{AhnMarron2010}.

\cite{MarronEtAl2007} viewed ``data-piling" as a drawback of the SVM, because the SVM classifier \eqref{eq:soln_SVM} is a function of only support vectors. Another popular classifier logistic regression does classification by using all the data points. However, the classical logistic regression
classifier is derived by following the maximum likelihood principle, not based on a nice margin-maximization motivation\footnote{\cite{ZhuHastie2005} later showed that the limiting $\ell_2$ penalized logistic regression approaches the margin-maximizing hyperplane for the separable data case. DWD was first proposed in 2002.}.
\cite{MarronEtAl2007} wanted to have a new method that is directly formulated by a SVM-like margin-maximization picture and also uses all data points for classification. To this end, \cite{MarronEtAl2007} proposed DWD which finds a separating hyperplane minimizing the total inverse margins of all the data points:
\begin{equation}
\begin{aligned}
\min_{\omega_0, \bs\omega}&\;\;\;  \left[\sum_{i=1}^n \dfrac{1}{d_i} + c \sum_{i=1}^n \eta_i \right],\\
\text{ subject to }&\;\;\; d_i = y_i(\omega_0 + \bs x_i^T\bs\omega) + \eta_i \ge 0, \ \eta_i \ge 0,\ \forall i, \text{ and } \bs\omega^T \bs\omega=1.
\end{aligned}
\label{eq:nonsep_DWD}
\end{equation}

There has been much work on variants of the standard DWD. We can only give an incomplete list here. \cite{QiaoEtAl2010} introduced the weighted DWD to tackle unequal cost or sample sizes by imposing different weights on two classes. \cite{HuangEtAl2013} extended the binary DWD to the multiclass case. \cite{WangZou2015} proposed the sparse DWD for high-dimensional classification. In addition, the work connecting DWD with other classifiers, e.g., SVM, includes but not limited to LUM \citep{LiuEtAl2011}, DWSVM \citep{QiaoZhang2015a}, and FLAME \citep{QiaoZhang2015b}. \cite{Marron2015} provided a more comprehensive review of the current DWD literature.

\subsubsection{Computation}
\cite{MarronEtAl2007} solved the standard DWD by reformulating \eqref{eq:nonsep_DWD} as a second-order cone programming (SOCP) program \citep{AlizadehGoldfarb2004, BoydVandenberghe2004}, which has a linear objective, linear constraints, and second-order-cone constraints. Specifically, for each $i$, let $\rho_i = (1/d_i + d_i)/2$, $\sigma_i = (1/d_i - d_i)/2$, and then $\rho_i + \sigma_i = 1/d_i$, $\rho_i - \sigma_i = d_i$, and $\rho_i^2 - \sigma_i^2 = 1$. Hence the original optimization problem \eqref{eq:nonsep_DWD} becomes
\begin{equation}
\begin{aligned}
\min_{\omega_0, \bs\omega} &\;\;\; \bigg[\bs 1^T \bs\rho + \bs 1^T \bs\sigma + c\bs 1^T \bs \eta \bigg],\\
\text{ subject to }&\;\;\;
\bs\rho - \bs\sigma = \tilde{\bs Y} \bs X \bs \omega + \omega_0 \cdot \bs y + \bs \eta, \\
&\;\;\; \eta_i \ge 0, \ (\rho_i; \sigma_i, 1) \in S_3, \ \forall i, \ (1; \bs\omega)\in S_{p+1},
\end{aligned}
\label{eq:primal_dwd}
\end{equation}
where $\tilde{\bs Y}$ is an $n \times n$ diagonal matrix with the $i$th diagonal element $y_i$, $\bs X$ is an $n \times p$ data matrix with the $i$th row $\bs x_i^T$, and $S_{m+1} = \{(\psi, \bs\phi)\in\mathbb{R}^{m+1}: \psi^2\ge \bs\phi^T \bs\phi \}$ is the form of the second-order cones.
After solving $\hat{\omega}_0$ and $\hat{\bs\omega}$ from \eqref{eq:primal_dwd}, a new observation $\bs x_{\mathrm{new}}$ is classified by $\mathrm{sign}(\hat{\omega_0} + \bs x_{\mathrm{new}}^T \hat{\bs\omega})$.

\subsubsection{Non-linear extension}
Note that the kernel SVM was derived from applying the kernel trick to the dual formulation (\ref{eq:soln_SVM}).
\cite{MarronEtAl2007} followed the same approach to consider a version of kernel DWD for achieving non-linear classification.
The dual function of the problem \eqref{eq:primal_dwd} is \citep{MarronEtAl2007}
\begin{equation}
\begin{aligned}
\max_{\bs \alpha} &\;\;\; \bigg[ - \sqrt{\bs\alpha^T \tilde{\bs Y} \bs X \bs X^T \tilde{\bs Y} \bs \alpha} + 2\cdot \bs 1^T \sqrt{\bs \alpha} \bigg],\\
\text{ subject to }&\;\;\; \bs y^T \bs \alpha = 0, \ \bs 0 \le \bs \alpha \le c\cdot \bs 1,
\end{aligned}
\label{eq:dual_dwd}
\end{equation}
where $(\sqrt{\bs \alpha})_i=\sqrt{\alpha_i}, \ i=1,2,\ldots,n$.
Note that \eqref{eq:dual_dwd} only uses $\bs X \bs X^T$, which makes it easy to employ the kernel trick to get a nonlinear extension of the linear DWD. For a given kernel function $K$, define the kernel matrix as $(\bs K)_{ij}=K(X_i,X_j)$, $1 \le i, \ j \le n$. Then a kernel DWD can be defined as \citep{MarronEtAl2007}
\begin{equation}
\begin{aligned}
\max_{\bs \alpha} &\;\;\; \bigg[ - \sqrt{\bs\alpha^T \tilde{\bs Y} \bs K \tilde{\bs Y} \bs \alpha} + 2\cdot \bs 1^T \sqrt{\bs \alpha} \bigg],\\
\text{ subject to }&\;\;\; \bs y^T \bs \alpha = 0, \ \bs 0 \le \bs \alpha \le c\cdot \bs 1.
\end{aligned}
\label{eq:dual_kerdwd}
\end{equation}
To solve (\ref{eq:dual_kerdwd}), \cite{MarronEtAl2007}  used the Cholesky decomposition of the kernel matrix, i.e., $\bs K = \bs \Phi \bs \Phi^T$ and then replaced the predictors $\bs X$ in \eqref{eq:primal_dwd} with $\bs \Phi$. \cite{MarronEtAl2007} also carefully discussed several algorithmic issues that ensure the equivalent optimality in \eqref{eq:primal_dwd} and \eqref{eq:dual_dwd}.

\vspace{0.1in}
\noindent {\bf Remark 1.} Two DWD implementations have been published thus far: a Matlab software \citep{Marron2013} and an R package \texttt{DWD} \citep{HuangEtAl2012}. Both implementations are based on a Matlab SOCP solver \texttt{SDPT3}, which was developed by \cite{TutuncuEtAl2003}. We notice that the R package \texttt{DWD} can only compute the linear DWD.

\vspace{0.1in}
\noindent {\bf Remark 2.} To our best knowledge, the theoretical justification for the kernel DWD in \cite{MarronEtAl2007} is still unclear. The reason is likely due to the fact that the nonlinear extension is purely algorithmic. 
In fact, the Bayes risk consistency of DWD was proposed as an open research problem in \cite{MarronEtAl2007}.
The kernel DWD considered in this paper can be rigorously justified to have a universal Bayes risk consistency property; see details in section~\ref{sec:learningtheory0}. 

\subsubsection{Generalized DWD}
\cite{MarronEtAl2007} also attempted to replace the reciprocal in the DWD optimization problem \eqref{eq:nonsep_DWD} with the $q$th power ($q > 0$) of the inverse distances, and \cite{HallEtAl2005} also used it as the original definition of DWD. We name the DWD with this new formulation the generalized DWD:
\begin{equation}
\begin{aligned}
\min_{\omega_0, \bs\omega}&\;\;\;  \left[\sum_{i=1}^n \dfrac{1}{d_i^q} + c \sum_{i=1}^n \eta_i \right],\\
\text{ subject to }&\;\;\; d_i = y_i(\omega_0 + \bs x_i^T\bs\omega) + \eta_i \ge 0, \ \eta_i \ge 0, \ \forall i, \text{ and } \bs\omega^T \bs\omega=1,
\label{eq:genDWD}
\end{aligned}
\end{equation}
which degenerates to the standard DWD \eqref{eq:nonsep_DWD} when $q=1$.


The first asymptotic theory for DWD and generalized DWD was given in \cite{HallEtAl2005} who presented a novel geometric representation of the HDLSS data. 
Assuming $\bs X_1^+, \bs X_2^+, \ldots, \bs X_{n^+}^+$ are the data from the positive class and $\bs X_1^-, \bs X_2^-, \ldots, \bs X_{n^-}^-$ are from the negative class.
\cite{HallEtAl2005} stated that, when the sample size $n$ is fixed and the dimension $p$ goes to infinity, under some regularity conditions, 
there exist two constants $l^+$ and $l^-$ such that for each pair of $i$ and $j$,
$$
\begin{aligned}
p^{-1/2}||\bs X_i^+ - \bs X_j^+|| \stackrel{P}{\to} \sqrt{2}l^+, \text{ and }
p^{-1/2}||\bs X_i^- - \bs X_j^-|| \stackrel{P}{\to} \sqrt{2}l^-,
\end{aligned}
$$
as $p \to \infty$. This result was applied the results to study several classifiers including the SVM and the generalized DWD. 
For ease presentation let us consider the equal subgroup size case, i.e., $n_+ = n_- = n/2$.
\cite{HallEtAl2005} assumed that
$
p^{-1/2} ||E\bs X^+ - E\bs X^-|| \to \mu, \text{ as } p \to \infty,
$ The basic conclusion is that
when $\mu$ is greater than a threshold that depends on $l^+,l^-,n$, the misclassification error converges to zero, and when $\mu$ is less than the same threshold, the misclassification error converges to $50\%$.
For more details, see Theorem 1 and Theorem 2 in \cite{HallEtAl2005}. \cite{AhnEtAl2007} further relaxed the assumptions thereof.

\vspace{0.1in}
\noindent {\bf Remark 3.} The generalized DWD has not been implemented yet because the SOCP transformation only works for the standard DWD ($q=1$) \eqref{eq:primal_dwd}, but its extension to handle the general cases is unclear if not impossible. That is why the current DWD literature only focuses on DWD with $q=1$. In fact, the generalized DWD with $q \neq 1$ was proposed as an open research problem in \cite{MarronEtAl2007}.
The new algorithm proposed in this paper can easily solve the generalized DWD problem for any $q>0$; see section~\ref{sec:computation}.

\section{A Novel Algorithm for DWD}
\label{sec:computation}
\cite{MarronEtAl2007} originally solved the standard DWD by transforming \eqref{eq:nonsep_DWD} into a SOCP problem. This algorithm, however, cannot compute the generalized DWD \eqref{eq:genDWD} with $q \neq 1$. In this section, we propose an entirely different algorithm based on the majorization-minimization (MM) principle. Our new algorithm offers a unified solution to the standard DWD and the generalized DWD.

\subsection{Generalized DWD loss}
\label{sec:dwdloss}
Our algorithm begins with a $loss+penalty$ formulation of the DWD. Lemma~\ref{lm:DWD_loss} deploys the result. Note that the loss function also lays the foundation of the kernel DWD learning theory that will be discussed in section~\ref{sec:kerDWD}.

\begin{Lemma}
The generalized DWD classifier in \eqref{eq:genDWD} can be written as $\mathrm{sign}(\hat{\beta}_0 + \bs x_i^T \hat{\bs\beta})$, where $(\hat{\beta}_0, \hat{\bs\beta})$ is computed from
\begin{equation}
\min_{\beta_0, \bs\beta}\bs C(\beta_0, \bs\beta) \equiv \min_{\beta_0, \bs\beta} \left[ \frac{1}{n} \sum_{i=1}^n V_q \left(y_i(\beta_0 + \bs x_i^T \bs\beta)\right) + \lambda \bs\beta^T \bs\beta \right],
\label{eq:DWDlossopt}
\end{equation}
for some $\lambda$, where
\begin{equation}
V_q(u)=
\begin{cases}
1-u, \; &\text{ if } u \le  \dfrac{q}{q+1},\\
      \dfrac{1}{{u}^q}\dfrac{q^q}{{(q+1)}^{q+1}}, \; &\text{ if } u > \dfrac{q}{q+1}.
    \end{cases}
\label{eq:DWDloss}
\end{equation}
\label{lm:DWD_loss}
\end{Lemma}

\vspace{0.1in}
\noindent {\bf Remark 4.} The proof of Lemma 1 provides the one-to-one mapping between $\lambda$ in \eqref{eq:DWDlossopt} and $c$ in \eqref{eq:genDWD}.
Write $(\hat{\beta}(\lambda)_0,\hat{\bs \beta}(\lambda))$ as the solution to \eqref{eq:DWDlossopt}. Define
$$
c(\lambda)=\frac{(q+1)^{q+1}}{q^q} \Vert \hat{\bs \beta}(\lambda)\Vert^{q+1}.
$$
Considering \eqref{eq:genDWD} using $c(\lambda)$,
\begin{equation}
\begin{aligned}
&(\hat{\omega_0},\hat{\bs \omega}) = \argmin_{\omega_0, \bs\omega}\;\;\;  \left[\sum_{i=1}^n \dfrac{1}{d_i^q} + c(\lambda) \sum_{i=1}^n \eta_i \right],\\
\text{ subject to }&\; d_i = y_i(\omega_0 + \bs x_i^T\bs\omega) + \eta_i \ge 0, \ \eta_i \ge 0, \ \forall i, \text{ and } \bs\omega^T \bs\omega=1,
\label{eq:genDWD2}
\end{aligned}
\end{equation}
we have
$$
\hat{\bs \omega}=\hat{\bs \beta}(\lambda)/\Vert \hat{\bs \beta}(\lambda)\Vert \text{ and }\hat{\omega}_0=\hat{\beta}(\lambda)_0/\Vert \hat{\bs \beta}(\lambda)\Vert.
$$
Note that $\mathrm{sign}(\hat{\omega}_0 + \bs x_i^T \hat{\bs\omega})=\mathrm{sign}(\hat{\beta}(\lambda)_0 + \bs x_i^T \hat{\bs\beta}(\lambda))$, which means that the generalized DWD classifier defined by \eqref{eq:genDWD2} is equivalent to the generalized DWD classifier defined by \eqref{eq:DWDlossopt}.

\begin{figure}[ht]
 \centering\includegraphics[width=.68\textwidth]{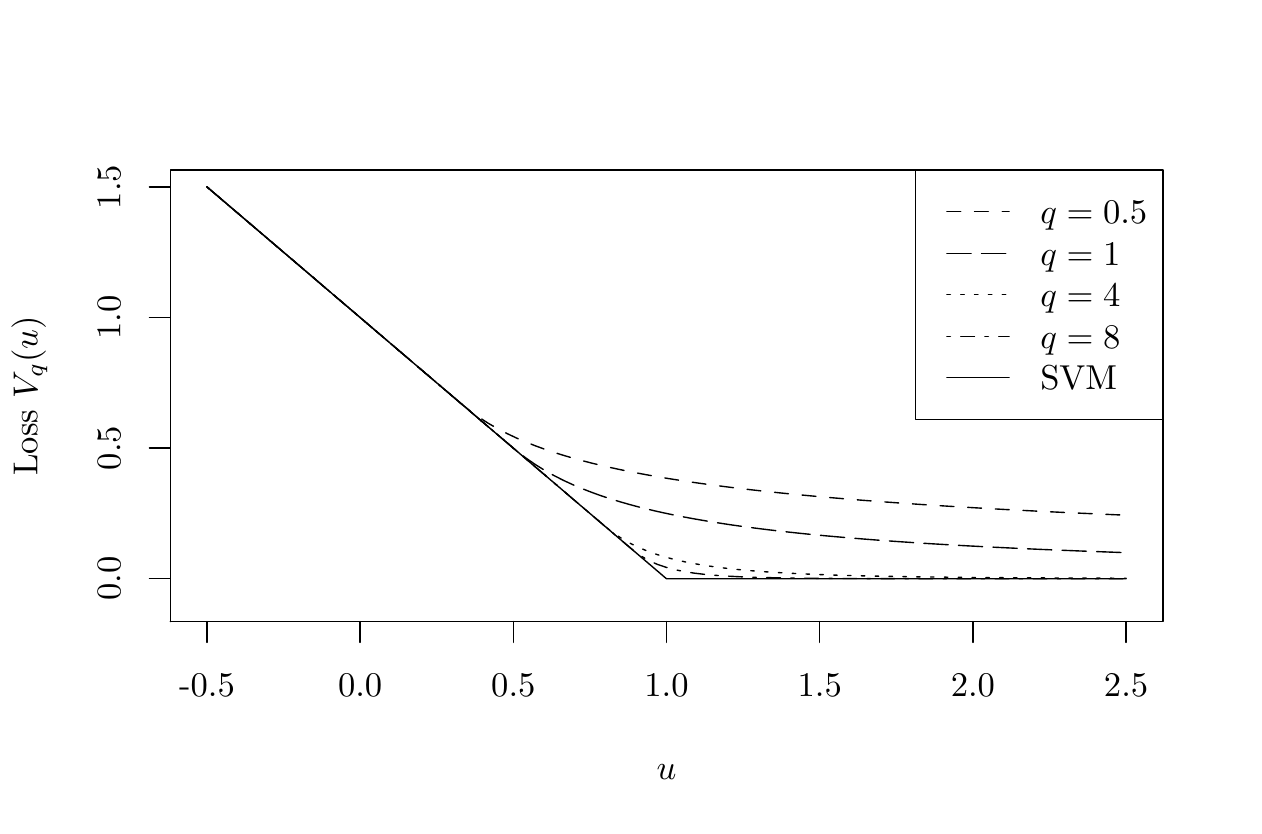}
 \caption{Top to bottom are the DWD loss functions with $q=0.5, 1, 4, 8$, and the SVM hinge loss. }
 \label{fig:dwdloss}
\end{figure}

By Lemma~\ref{lm:DWD_loss}, we call $V_q(\cdot)$ the generalized DWD loss. It can be visualized in Figure~\ref{fig:dwdloss}. We observe that the generalized DWD loss decreases as $q$ increases and it approaches the SVM hinge loss function as $q \to \infty$.
When $q=1$, the generalized DWD loss becomes
\begin{equation*}
V_1(u)=
\begin{cases}
\setstretch{1.11}
1-u, \; &\text{ if } u \le  1/2,\\
      1/(4u), \; &\text{ if } u > 1/2.
    \end{cases}
\end{equation*}
We notice that $V_1(u)$ has appeared in the literature \citep{QiaoEtAl2010, LiuEtAl2011}. In this work we give a unified treatment of all $q$ values, not just $q=1$.


\subsection{Derivation of the algorithm}
\label{sec:deviation}
We now show how to develop the new algorithm by using the MM principle \citep{DeLeeuwHeiser1977, LangeEtAl2000, HunterLange2004}. Some recent successful applications of the MM principle can be seen in \cite{HunterLi2005, WuLange2008, ZouLi2008, ZhouLange2010, YangZou2013, LangeZhou2014}, among others. The main idea of the MM principle is easy to understand. Suppose $\bs\theta=(\beta_0, \bs\beta^T)^T$ and we aim to minimize $\bs C(\bs\theta)$, defined in \eqref{eq:DWDlossopt}. The MM principle finds a majorization function $\bs D(\bs\theta| \bs\theta_{k})$ satisfying $\bs C(\bs\theta) < \bs D(\bs\theta| \bs\theta_{k})$ for any $\bs\theta \neq \bs\theta_{k}$ and $\bs C( \bs\theta_{k}) = \bs D(\bs\theta_{k}|\bs\theta_{k})$, and then we generate a sequence $\{\bs C( \bs\theta_{k})\}_{k=1}^\infty$ by updating $\bs\theta_{k}$ via $\bs\theta_{k} \leftarrow  \bs\theta_{k+1} = \argmin_{\bs\theta} \bs D(\bs\theta| \bs\theta_{k})$.

We first expose some properties of the generalized DWD loss functions, which give rise to a quadratic majorization function of $\bs C(\bs\theta)$.
The generalized DWD loss is differentiable everywhere; its first-order derivative is given below,
\begin{equation}
V_q'(u)=
\begin{cases}
-1, \; &\text{ if } u \le  \dfrac{q}{q+1},\\
-\dfrac{1}{u^{q+1}}{\left(\dfrac{q}{q+1}\right)}^{q+1},  &\text{ if } u > \dfrac{q}{q+1}.
\end{cases}
\end{equation}

\begin{Lemma} 
The generalized DWD loss function $V_q(\cdot)$ has a Lipschitz continuous gradient,
\begin{equation}
|V'_q(t) - V'_q(\tilde{t})| < M|t - \tilde{t}|,
\label{eq:Lip_condtn}
\end{equation}
which further implies a quadratic majorization function of $V_q(\cdot)$ such that
\begin{equation}
V_q(t) < V_q(\tilde{t}) + V'_q(\tilde{t})(t-\tilde{t}) + \frac{M}{2}(t-\tilde{t})^2
\label{eq:quad_condtn}
\end{equation}
for any $t \neq \tilde{t}$ and $M=(q+1)^2/q$.
\label{lm:Lps}
\end{Lemma}
Denote the current solution by $\tilde{\bs\theta}=(\tilde{\beta}_0, \tilde{\bs\beta}^T)^T$ and the updated solution by $\bs\theta=(\beta_0, \bs\beta^T)^T$. We settle $\bs C({\bs\theta})=\bs C({\beta}_0, {\bs\beta})$ and $\bs D(\bs\theta|\tilde{\bs\theta})=\bs D(\beta_0, \bs\beta)$ without abusing notations. We have that for any $(\beta_0, \bs\beta)\neq (\tilde{\beta}_0, \tilde{\bs\beta})$,
\begin{equation}
\begin{split}
&\bs C(\beta_0, \bs\beta)\\
\equiv&\frac{1}{n} \sum_{i=1}^n V_q\left(y_i(\beta_0 + \bs x_i ^T \bs\beta)\right)
+ \lambda \bs \beta^T \bs \beta \\
< & \frac{1}{n} \sum_{i=1}^n V_q\left(y_i(\tilde{\beta}_0 + \bs x_i ^T \tilde{\bs\beta})\right)
+ \frac{1}{n} \sum_{i=1}^n V_q'\left(y_i(\tilde{\beta}_0 + \bs x_i ^T \tilde{\bs\beta})\right)\left[y_i(\beta_0 - \tilde\beta_0) + y_i \bs x_i^T (\bs\beta - \tilde{\bs\beta})\right] \\
&+ \frac{M}{2n} \sum_{i=1}^n \left[y_i(\beta_0 - \tilde\beta_0) + y_i \bs x_i^T (\bs\beta - \tilde{\bs\beta})\right]^2
+ \lambda \bs\beta^T \bs\beta \\
\equiv & \bs D(\beta_0, \bs\beta).
\label{eq:CleD}
\end{split}
\end{equation}

We now find the minimizer of  $\bs D(\beta_0, \bs\beta)$.
The gradients of  $\bs D(\beta_0, \bs\beta)$ are given as follows:
\begin{align}
\partial\frac{\bs D(\beta_0, \bs\beta)}{\partial \bs\beta}
=& \frac{1}{n} \sum_{i=1}^n V_q'\left(y_i(\tilde{\beta}_0 + \bs x_i ^T \tilde{\bs\beta})\right) y_i \bs x_i + \frac{M}{n}\sum_{i=1}^n \left[(\beta_0 - \tilde\beta_0) + \bs x_i^T (\bs\beta - \tilde{\bs\beta})\right]\bs x_i + 2\lambda \bs\beta \notag\\
=& \bs X^T \bs z + \frac{M}{n}(\beta_0 - \tilde{\beta}_0)\bs X^T \bs 1 + \frac{M}{n}\sum_{i=1}^n \bs x_i \bs x_i^T (\bs\beta - \tilde{\bs\beta}) + 2\lambda \bs\beta \notag\\
=& \bs X^T \bs z + \frac{M}{n}(\beta_0 - \tilde{\beta}_0)\bs X^T \bs 1 + \left(\frac{M}{n} \bs X^T \bs X + 2\lambda\bs I_p\right) (\bs\beta - \tilde{\bs\beta}) + 2\lambda \tilde{\bs\beta}\label{eq:grdnt_alp}, \\
\partial\frac{\bs D(\beta_0, \bs\beta)}{\partial \beta_0}
=& \frac{1}{n} \sum_{i=1}^n V_q'\left(y_i(\tilde{\beta}_0 + \bs x_i ^T \tilde{\bs\beta})\right) y_i + \frac{M}{n} \sum_{i=1}^n \left[(\beta_0 - \tilde{\beta}_0) + \bs x_i^T(\bs\beta - \tilde{\bs\beta})\right]  \notag\\
=& \bs 1^T \bs z + M(\beta_0 - \tilde{\beta}_0) + \frac{M}{n} \bs 1^T \bs X (\bs\beta - \tilde{\bs\beta}).
\label{eq:grdnt_b0}
\end{align}
where $\bs X$ is the $n \times p$ data matrix with the $i$th row $\bs x_i^T$, $\bs z$ is an $n \times 1$ vector with the $i$th element $y_i V_q'(y_i (\tilde{\beta}_0+\bs x_i^T \tilde{\bs\beta}))/n$, and $\bs 1\in \mathbb{R}^n$ is the vector of ones. Setting $[\partial\bs D(\beta_0, \bs\beta)/\partial \beta_0, \partial\bs D(\beta_0, \bs\beta)/\partial \bs\beta]$ to be zeros, we obtain the minimizer of $\bs D(\beta_0, \bs\beta)$:
\begin{equation}
\begin{aligned}
\left( \begin{array}{c}
\beta_0\\
\bs\beta
 \end{array} \right)
 =
 \left( \begin{array}{c}
\tilde{\beta}_0 \\
\tilde{\bs\beta}
 \end{array} \right)
- \frac{n}{M}
\left( \begin{array}{cc}
n & \bs 1^T \bs X  \\
\bs X^T \bs 1 & \bs X^T \bs X + \frac{2n\lambda}{M} \bs I_p
 \end{array} \right)^{-1}
\left( \begin{array}{c}
 \bs 1^T \bs z \\
 \bs X^T \bs z + 2\lambda \tilde{\bs\beta}
 \end{array} \right).
\end{aligned}
\label{eq:updt}
\end{equation}
So far we have completed all the steps of the MM algorithm. Details are summarized in Algorithm~\ref{alg:linear}.

\begin{algorithm}[t]
\begin{algorithmic}[1]
\caption{Linear generalized DWD
\label{alg:linear}}
\STATE Initialize $(\tilde{\beta}_0, \tilde{\bs\beta}^T)$
\FOR {each $\lambda$}
\STATE Compute $\bs P^{-1}(\lambda)$:
\begin{align*}
\bs P^{-1}(\lambda) =
\left( \begin{array}{cc}
n & \bs 1^T \bs X  \\
\bs X^T \bs 1 & \bs X^T \bs X + \frac{2n\lambda}{M} \bs I_p
 \end{array} \right)^{-1}
\end{align*}
\REPEAT {}
\STATE Compute $\bs z =(z_1, \ldots, z_n)^T$: $z_i =y_i V_q'(y_i (\tilde{\beta}_0+\bs x_i \tilde{\bs\beta}))/n$
\STATE Compute:
\begin{align*}
\left( \begin{array}{c}
\beta_0 \\
\bs\beta
 \end{array} \right)
 \leftarrow
\left( \begin{array}{c}
\tilde{\beta}_0 \\
\tilde{\bs\beta}
 \end{array} \right)
- \frac{nq}{(q+1)^2}
\bs P^{-1}(\lambda)
\left( \begin{array}{c}
 \bs 1^T \bs z \\
 \bs X^T \bs z + 2\lambda \tilde{\bs\beta}
 \end{array} \right)
\end{align*}
\STATE Set $(\tilde{\beta}_0, \tilde{\bs\beta}^T)$ = $(\beta_0, \bs\beta^T)$
\UNTIL {the convergence condition is met}
\ENDFOR
\end{algorithmic}
\end{algorithm}

We have implemented Algorithm~\ref{alg:linear} in an R package \texttt{kerndwd}, which is publicly available for download on CRAN.


\subsection{Performance of the new algorithm}
\label{sec:performance}
In this section, we show the superior computation performance of our R implementation, \texttt{kerndwd}, over the two existing implementations, the R package \texttt{DWD} \citep{HuangEtAl2012} and the Matlab software \citep{Marron2013}. To avoid confusion, we henceforth use \texttt{OURS}, \texttt{HUANG}, and \texttt{MARRON} to denote \texttt{kerndwd}, \texttt{DWD}, and the Matlab implementation, respectively. Since \texttt{HUANG} is incapable of non-linear kernels and the generalized DWD with $q\neq 1$, we only attend to the linear DWD with $q$ fixed to be one. All experiments were conducted on an Intel Core i5 M560 (2.67 GHz) processor.

For a fair comparison, we study the four numerical examples used in \cite{MarronEtAl2007}, except for different sample sizes and dimensions. In each example, we generate a data set with sample size $n=500$ and dimension $p=50$. The responses are always binary; one half of the data have responses $+1$ and the other half have $-1$. Data in example 1 are generated from Gaussian distribution with means of $(\pm 2.2, 0, \ldots, 0)$ and an identity covariance for $\pm 1$ classes respectively. Example 2 has 80\% of data drawn as example 1 whereas the other 20\% from Gaussian distributions with means of $(\pm 100, \pm 500, 0, \ldots, 0)$ for $\pm 1$ classes. In example 3, 80\% of the data are obtained as example 1 as well, while the means of the remaining 20\% have the first coordinate replaced by $\pm 0.1$ and one randomly chosen coordinate replaced by $\pm 100$ for $\pm 1$ classes. For example 4, at the first 25 coordinates, the data from $-1$ class are standard Gaussian and the data from $+1$ class are $11.09$ times standard Gaussian; for both classes, the last 25 coordinates are just the squares of the first 25.

In each example, we fitted a linear DWD with five different tuning parameter values $\lambda=(0.01, 0.1, 1, 10, 100)$.
After obtaining $(\hat \beta_0,\hat {\bs \beta})$, we computed $(\hat {\omega}_0,\hat {\bs \omega}$) and the constant $c$ in \eqref{eq:primal_dwd} by using Remark 4. We then used \texttt{HUANG} and \texttt{MARRON} to compute their solutions. Note that in theory all three implementations should yield identical  $(\hat {\omega}_0,\hat {\bs \omega}).$ From table~\ref{tab:simu} we observe that \texttt{OURS} took remarkably less computation time than \texttt{HUANG} and \texttt{MARRON}. In example 1, for instance, \texttt{OURS} spent only 0.012 second on average to fit a DWD model, while \texttt{HUANG} used 14.525 seconds, and \texttt{MARRON} took 2.204 seconds, which were 1210 and 183 times larger, respectively. In all four examples, the timings of \texttt{OURS} were 700 times above faster than the existing R implementation \texttt{HUANG}, and also more than 70 times faster than the Matlab implementation \texttt{MARRON}\footnote{We also checked the quality of the computed solutions by these different algorithms. In theory they should be identical. In practice, due to machine errors and implementations, they could be different. We found that in all examples our new algorithm gave better solutions in the sense that the objective function in \eqref{eq:primal_dwd} has the smallest value. \texttt{HUANG} and \texttt{MARRON} gave similar but slightly larger objective function values. }.

\begin{table}[ht]
\centering
\caption{Timing comparisons among the R package \texttt{kerndwd} (denoted as \texttt{OURS}), the R package \texttt{DWD} (denoted as \texttt{HUANG}), and the Matlab implementation (denoted as \texttt{MARRON}). All the timings are averaged over 100 independent replicates.}
\resizebox{\textwidth}{!}{\begin{minipage}{1.2\textwidth}
\label{tab:simu}
\ra{1.42}
\centering
\begin{tabular}{@{}ccccccccccccccccccc}\toprule
&&\multicolumn{3}{c}{\texttt{Timing (in sec.)}} & &\multicolumn{2}{c}{\texttt{Ratio}}  \\
\cmidrule{3-5} \cmidrule{7-8}
&& \multicolumn{1}{c}{\texttt{OURS}} &\multicolumn{1}{c}{\texttt{HUANG}} &\multicolumn{1}{c}{\texttt{MARRON}} &&\multicolumn{1}{c}{\small $\dfrac{t(\texttt{HUANG})}{t(\texttt{OURS})}$} & \multicolumn{1}{c}{\small $\dfrac{t(\texttt{MARRON})}{t(\texttt{OURS})}$}\\
\midrule
1 && 0.012 & 14.525 & 2.204 & & 1210.8 & 183.7 \\
2 && 0.024 & 18.018 & 2.411 & & 750.8 & 100.5 \\
3 && 0.028 & 26.918 & 2.076 & & 961.4 & 74.1 \\
4 && 0.020 & 21.536 & 2.264 & & 1076.8 & 113.2 \\
\bottomrule
\end{tabular}
 \end{minipage}}
\end{table}

\section{Kernel DWD in RKHS and Bayes Risk Consistency}\label{sec:kerDWD}
\subsection{Kernel DWD in RKHS}
\label{sec:RKHS}
The kernel SVM can be derived by using the kernel trick or using the view of non-parametric function estimation in a reproducing kernel Hilbert space (RKHS). Much of the theoretical work on the kernel SVM is based on the RKHS formulation of SVMs. The derivation of the kernel SVM in a RKHS is given in \cite{HastieEtAl2009}. We take a similar approach to derive the kernel DWD, as our goal is to establish the kernel learning theory for DWD.

Consider $\mathcal{H}_K$, a reproducing kernel Hilbert space generated by the kernel function $K$. The Mercer's theorem ensures $K$ to have an eigen-expansion $K(\bs x, \bs x') = \sum_{t=1}^\infty \gamma_t \phi_t(\bs x) \phi^T_t(\bs x')$, with $\gamma_t \ge 0$ and $\sum_{t=1}^\infty \gamma_t^2 < \infty$. Then the Hilbert space $\mathcal{H}_K$ is defined as the collection of functions $h(\bs x) = \sum_{t=1}^\infty \theta_t\phi_t(\bs x)$, for any $\theta_t$ such that $\sum_{t=1}^\infty \theta_t^2/\gamma_t<\infty$, and the inner product is $\left\langle \sum_{t=1}^\infty \theta_t\phi_t(\bs x), \sum_{t'=1}^\infty \delta_{t'}\phi_{t'}(\bs x) \right\rangle _{\mathcal{H}_K} = \sum_{t=1}^\infty \theta_t \delta_t/\gamma_t$.

Given $\mathcal{H}_K$, let the non-linear DWD be written as $\mathrm{sign}(\hat{\beta}_0 + \hat{h}(\bs x))$ where $(\hat{\beta}_0, \hat{h})$ is the solution of
\begin{equation}
\min_{\substack{h \in \mathcal{H}_K \\ \beta_0 \in \mathbb{R}}} \left[ \dfrac{1}{n}\sum_{i=1}^n V_q \left(y_i (\beta_0 + h(\bs x_i)) \right) + \lambda||h||^2_{\mathcal{H}_K} \right],
\label{eq:nonlnr_DWD1}
\end{equation}
where $V_q(\cdot)$ is the generalized DWD loss \eqref{eq:DWDloss}.
The representer theorem concludes that the solution of \eqref{eq:nonlnr_DWD1} has a finite expansion based on $K(\bs x, \bs x_i)$ \citep{Wahba1990},
$$
\hat{h}(\bs x) = \sum_{i=1}^n \hat{\alpha}_i K(\bs x, \bs x_i),
$$
and thus
$$
||\hat{h}||^2_{\mathcal{H}_K} = \sum_{i=1}^n \sum_{j=1}^n \hat{\alpha}_i \hat{\alpha}_j K(\bs x_i, \bs x_j).
$$
Consequently, \eqref{eq:nonlnr_DWD1} can be paraphrased with matrix notation,
\begin{equation}
\min_{\beta_0, \bs\alpha}\bs C_K(\beta_0, \bs\alpha) \equiv \min_{\beta_0, \bs\alpha} \left[ \frac{1}{n} \sum_{i=1}^n V_q \left( y_i(\beta_0 + \bs K_i^T \bs\alpha) \right) + \lambda \bs \alpha^T \bs K \bs \alpha \right],
\label{eq:nonlnr_DWD}
\end{equation}
where $\bs K$ is the kernel matrix with the $(i, j)$th element of $K(\bs x_i, \bs x_j)$ and $\bs K_i$ is the $i$th column of $\bs K$.

\vspace{0.1in}
\noindent {\bf Remark 5.} We can compare (\ref{eq:nonlnr_DWD}) to the kernel SVM \citep{HastieEtAl2009}
\begin{equation}
\min_{\beta_0, \bs\alpha} \left[ \frac{1}{n} \sum_{i=1}^n \left[1 - y_i(\beta_0 + \bs K_i^T \bs\alpha) \right]_+ + \lambda \bs \alpha^T \bs K \bs \alpha \right],
\label{eq:optmztn}
\end{equation}
where $[1-t]_+$ is the hinge loss underlying the SVM. As shown in Figure 3, the generalized DWD loss takes the hinge loss as its limit when $q \rightarrow \infty$. In general, the generalized DWD loss and the hinge loss look very similar, which suggests that the kernel DWD and the kernel SVM equipped with the same kernel have similar statistical behavior.

The procedure for deriving Algorithm~\ref{alg:linear} for the linear DWD can be directly adopted to derive an efficient algorithm
for solving the kernel DWD. We obtain the majorization function $\bs D_K(\beta_0, \bs\alpha)$,
\begin{eqnarray*}
\bs D_K(\beta_0, \bs\alpha) &=  & \frac{1}{n} \sum_{i=1}^n V_q'\left(y_i(\tilde{\beta}_0 + \bs K_i ^T \tilde{\bs\alpha})\right)\left[y_i(\beta_0 - \tilde\beta_0) + y_i \bs K_i^T (\bs\alpha - \tilde{\bs\alpha})\right]+\lambda \bs \alpha^T \bs K \bs \alpha \\
&&+ \frac{M}{2n} \sum_{i=1}^n \left[y_i(\beta_0 - \tilde\beta_0) + y_i \bs K_i^T (\bs\alpha - \tilde{\bs\alpha})\right]^2
+\frac{1}{n} \sum_{i=1}^n V_q\left(y_i(\tilde{\beta}_0 + \bs K_i ^T \tilde{\bs\alpha})\right) 
\end{eqnarray*}
and then find the minimizer of  $\bs D_K(\beta_0, \bs\alpha)$ which has a closed-form expression.
We opt to omit the details here for space consideration.  Algorithm~\ref{alg:kernel} summarizes the entire algorithm for the kernel DWD. 

\begin{algorithm}[t]
\begin{algorithmic}[1]
\caption{Kernel DWD
\label{alg:kernel}}
\STATE Initialize $(\tilde{\beta}_0, \tilde{\bs\alpha}^T)$
\FOR {each $\lambda$}
\STATE Compute $\bs P^{-1}(\lambda)$:
\begin{align*}
\bs P^{-1}(\lambda) =
\left( \begin{array}{cc}
n & \bs 1^T \bs K  \\
\bs K \bs 1 & \bs{KK} + \frac{2nq\lambda}{(q+1)^2} \bs K
 \end{array} \right)^{-1}
\end{align*}
\REPEAT {}
\STATE Compute $\bs z =(z_1, \ldots, z_n)^T$: $z_i =y_i V_q'(y_i (\tilde{\beta}_0+\bs K_i \tilde{\bs\alpha}))/n$
\STATE Compute:
\begin{align*}
\left( \begin{array}{c}
\beta_0 \\
\bs\alpha
 \end{array} \right)
 \leftarrow
\left( \begin{array}{c}
\tilde{\beta}_0 \\
\tilde{\bs\alpha}
 \end{array} \right)
- \frac{nq}{(q+1)^2}
\bs P^{-1}(\lambda)
\left( \begin{array}{c}
 \bs 1^T \bs z \\
 \bs{Kz} + 2\lambda \bs K \tilde{\bs\alpha}
 \end{array} \right)
\end{align*}
\STATE Set $(\tilde{\beta}_0, \tilde{\bs\alpha}^T)$ = $(\beta_0, \bs\alpha^T)$
\UNTIL {the convergence condition is met}
\ENDFOR
\end{algorithmic}
\end{algorithm}

\subsection{Kernel learning theory}
\label{sec:learningtheory0}
\cite{Lin2002} formulated the kernel SVM as a non-parametric function estimation problem in a reproducing kernel Hilbert space and
showed that the population minimizer of the SVM loss function is the Bayes rule, indicating that the SVM directly approximates the optimal Bayes classifier. 
\cite{Lin2004} further coined a name ``Fisher consistency" to describe such a result. The Vapnik-Chervonenkis (VC) analysis \citep{Vapnik1998, AnthonyBartlett1999} and the margin analysis \citep{BartlettShaweTaylor1999, ShaweTaylorCristianini2000} have been used to bound the expected classification error of the SVM.
\cite{Zhang2004} used the so-called leave-one-out analysis \citep{JaakkolaHaussler1999} to study a class of kernel machines.
The exisiting theoretical work on the kernel SVM provides us a nice road map to study the kernel DWD. 
In this section we first elucidate the Fisher consistency \citep{Lin2004} of the generalized kernel DWD, and we then establish the Bayes risk consistency of the kernel DWD when a universal kernel is employed.

Let $\eta(\bs x)$ denote the conditional probability $P(Y=1|\bs X=\bs x)$. Under the 0-1 loss, the theoretical optimal Bayes rule is $f^\star(\bs x) = \mathrm{sign}(\eta(\bs x)-1/2)$. Assume $\eta(\bs x)$ is a measurable function and $P(\eta(\bs x) = 1/2)=0$ throughout. 

\begin{Lemma}
The population minimizer of the expected generalized DWD loss $E_{\bs XY}[V_q\left(Yf(\bs X)\right)]$ is
\begin{equation}
\tilde{f}(\bs x) = \dfrac{q}{q+1} \left[                                                                                                \left( \dfrac{\eta(\bs x)}{1-\eta(\bs x)} \right)^{\frac{1}{q+1}}\cdot I(\eta(\bs x)>1/2) - \left( \dfrac{1-\eta(\bs x)}{\eta(\bs x)} \right)^{\frac{1}{q+1}}\cdot I(\eta(\bs x)<1/2) \right],
\label{eq:Fisher_min}
\end{equation}
where $I(\cdot)$ is the indicator function. The population minimizer $\tilde{f}(\bs x)$ has the same sign as $\eta(\bs x)-1/2$.
\label{lm:Fisher}
\end{Lemma}

\vspace{0.1in}
%

Fisher consistency is a property of the loss function. The interpretation is that the generalized DWD can approach Bayes rule with infinite many samples. 
We notice that Fisher consistency of $V_1(u)$ has been shown before \citep{QiaoEtAl2010, LiuEtAl2011}. 
In reality all classifiers are estimated from a finite sample. Thus, a more refined analysis of the actual DWD classifier is needed, and that is what we achieve in the following.


Following the convention in the literature, we absorb the intercept into $h$ and present the kernel DWD as follows:
\begin{equation}
\hat{f}_n = \argmin_{f \in {\mathcal{H}}_K} \left[ \dfrac{1}{n}\sum_{i=1}^n V_q \left( y_i (f(\bs x_i) \right) + \lambda_n||f||^2_{{\mathcal{H}}_K} \right].
\label{eq:ker_DWD}
\end{equation}
The ultimate goal is to show that the misclassification error of the kernel DWD approaches the Bayes error rate such that we can say the kernel DWD classifier works as well as the Bayes rule (asymptotically speaking). 
Following \cite{Zhang2004}, we derive the following lemma.

\begin{Lemma}
For a discrimination function $f$, we define
$
R(f)=E_{\bs XY}\left[Y \neq \mathrm{sign}\left( f(\bs X) \right) \right].
$
Assume that $f^\star = \argmin_f R(f)$ is the Bayes rule and $\hat{f}_n$ is the solution of \eqref{eq:ker_DWD}, then
\begin{equation}
R(\hat{f}_n) - R(f^\star) \le \dfrac{q+1}{q}(\varepsilon_A + \varepsilon_E),
\label{eq:err_fhat}
\end{equation}
where $\varepsilon_A$ and $\varepsilon_E$ are defined as follows and $V_q$ is the generalized DWD loss,
\begin{equation}
\begin{aligned}
\varepsilon_A &= \inf_{f\in{\mathcal{H}}_K} E_{\bs XY}\bigg[V_q(Yf(\bs X))\bigg] - E_{\bs XY}\bigg[V_q \left(Y\tilde{f}(\bs X)\right)\bigg],\\
\varepsilon_E=\varepsilon_E(\hat{f}_n) &= E_{\bs XY}\bigg[V_q \left(Y\hat{f}_n(\bs X)\right)\bigg] - \inf_{f\in{\mathcal{H}}_K} E_{\bs XY}\bigg[V_q(Yf(\bs X))\bigg].
\end{aligned}
\label{eq:two_err}
\end{equation}
\label{lm:err_bd}
\end{Lemma}

In the above lemma $R(f^*)$ is the Bayes error rate and $R(\hat{f}_n) $ is the misclassification error of the kernel DWD applied to new data points.
If $R(\hat{f}_n) \rightarrow R(f^\star) $, we say the classifier is Bayes risk consistent. Based on Lemma~\ref{lm:err_bd}, it suffices to show that both $\varepsilon_A$ and $\varepsilon_E$ approach zero in order to demonstrate the Bayes risk consistency of the kernel DWD. Note that $\varepsilon_A$ is deterministic and is called the approximation error. If the RKHS is rich enough then the approximation error can be made arbitrarily small. In the literature, the notation of \textit{universal kernel}  \citep{Steinwart2001,MicchelliEtAl2006} has been proposed and studied. Suppose $\mathcal{X}\in \mathbb{R}^p$ is the compact input space of $\bs X$ and $C(\mathcal{X})$ is the space of all continuous functions $g:\mathcal{X}\to\mathbb{R}$. The kernel $K$ is said to be \textit{universal} if the function space ${\mathcal{H}}_K$ generated by $K$ is dense in $\mathcal{C}(\mathcal{X})$, that is, for any positive $\epsilon$ and any function $g\in \mathcal{C}(\mathcal{X})$, there exists an $f \in {\mathcal{H}}_K$ such that $||f-g||_\infty < \epsilon$.

\begin{theorem}
Suppose $\hat{f}_n$ is the solution of \eqref{eq:ker_DWD}, ${\mathcal{H}}_K$ is induced by a universal kernel $K$, and the sample space $\mathcal{X}$ is compact.
Then we have 
\begin{itemize}
\item[(1)] $\varepsilon_A=0$;
\item[(2)]  
Let $B=\sup_{\bs x}K(\bs x,\bs x)<\infty$. When $\lambda_n \rightarrow 0$ and $n\lambda_n \rightarrow \infty$,  for any $\epsilon>0$, 
$$
\lim_{n \to \infty}P\left(  \varepsilon_E(\hat{f}_n)  > \epsilon \right) = 0.
$$
\end{itemize}
By (1) and (2) and \eqref{eq:err_fhat} we have $R(\hat f_n) \rightarrow R(f^*)$ in probability.
\label{lm:thm}
\end{theorem}

The Gaussian kernel is universal and $B \le 1$. Thus Theorem~\ref{lm:thm} says that the kernel DWD using the Gaussian kernel is Bayes risk consistent. This offers a theoretical explanation to the numerical results in Figure~\ref{fig:svm_dwd}.

\section{Real Data Analysis}
\label{sec:real_data}
In this section, we investigate the performance of \texttt{kerndwd} on four benchmark data sets: the BUPA liver disorder data, the Haberman's survival data, the Connectionist Bench (sonar, mines vs. rocks) data, and the vertebral column data. All the data sets were obtained from UCI Machine Learning Repository \citep{Lichman2013}.

For comparison purposes, we considered the SVM, the standard DWD ($q=1$) and the generalized DWD models with $q=0.5, 4, 8$. We computed  all DWD models using our R package \texttt{kerndwd} and solved the SVM using the R package \texttt{kernlab} \citep{KaratzoglouEtAl2004}. We randomly split each data into a training and a test set with a ratio $2:1$. For each method using the linear kernel, we conducted a five-folder cross-validation on the training set to tune $\lambda$.
For each method using Gaussian kernels, the pair of $(\sigma, \lambda)$ was tuned by the five-folder cross-validation.
We then fitted each model with the selected $\lambda$ and evaluated its prediction accuracy on the test set.

Table~\ref{tab:realdata} displays the average timing and mis-classification rates. We do not argue that either SVM or DWD outperforms the other; nevertheless, two models are highly comparable. SVM models work better on sonar and vertebral data, and DWD performs better on bupa and haberman data. For three out of the four data sets, the best method uses a Gaussian kernel, indicating that linear classifiers may not be adequate in such cases. In terms of timing, \texttt{kerndwd} runs faster than \texttt{kernlab} in all these examples. It is also interesting to see that DWD with $q=0.5$ can work slightly better than DWD with $q=1$ on bupa and haberman data, although the difference is not significant.

\begin{table}[ht]
\caption{The mis-classification rates and timings (in seconds) for four benchmark data sets. Each data set was split into a training and a test set. On the training set, the tuning parameters were selected by five-fold cross-validation and the models were fitted accordingly. The mis-classification rates were assessed on the test sets. All the timings include tuning parameters. For each dataset, the method with the best prediction accuracy is marked by black boxes.}
\resizebox{\textwidth}{!}{\begin{minipage}{1.39\textwidth}
\label{tab:realdata}
\ra{1.32}
\centering
\begin{tabular}{clrrrrrrrrrrrrrrrrr}
\toprule
&& \multicolumn{3}{c}{Bupa} &\phantom{} & \multicolumn{3}{c}{Haberman}&\phantom{}&
 \multicolumn{3}{c}{Sonar}&\phantom{}& \multicolumn{3}{c}{Vertebral}&\\
&& \multicolumn{3}{c}{\footnotesize{$n=345$, $p=6$}} &\phantom{} & \multicolumn{3}{c}{\footnotesize{$n=305$, $p=3$}}&\phantom{}
&\multicolumn{3}{c}{\footnotesize{$n=208$, $p=60$}}&\phantom{}& \multicolumn{3}{c}{\footnotesize{$n=310$, $p=6$}}\\
\cmidrule{3-5} \cmidrule{7-9} \cmidrule{11-13}  \cmidrule{15-17}
&&\multicolumn{2}{c}{error $(\%)$}  & time && \multicolumn{2}{c}{error $(\%)$}  & time && \multicolumn{2}{c}{error $(\%)$} & time && \multicolumn{2}{c}{error $(\%)$} & time
\\
  \midrule
\parbox[t]{2mm}{\multirow{5}{*}{\rotatebox[origin=c]{90}
{linear kernel}}}&SVM &  31.63 & (0.50) & 17.47 &  & 26.97 & (0.53) & 11.74 &  & 25.97 & (0.66) & 8.01 &  & \framebox{\textbf{14.83}} & (0.42) & 8.07 \\
&DWD $q=1$ & 34.82 & (0.75) & 0.05 &  & 26.71 & (0.54) & 0.03 &  & 25.65 & (0.75) & 0.30 &  & 16.76 & (0.53) & 0.07 \\
&DWD $q=0.5$& 34.23 & (0.72) & 0.06 &  & 26.73 & (0.53) & 0.04 &  & 25.10 & (0.72) & 0.35 &  & 16.54 & (0.51) & 0.10 \\
&DWD $q=4$& 35.08 & (0.71) & 0.05 &  & 26.69 & (0.55) & 0.03 &  & 26.00 & (0.76) & 0.32 &  & 16.54 & (0.53) & 0.06 \\
&DWD $q=8$ & 35.08 & (0.76) & 0.06 &  & 26.53 & (0.56) & 0.03 &  & 25.97 & (0.71) & 0.34 &  & 17.01 & (0.53) & 0.06 \\
\cmidrule{1-18}
\parbox[t]{2mm}{\multirow{5}{*}{\rotatebox[origin=c]{90}{Gaussian kernel}}}&SVM  & 32.23 & (0.48) & 6.57 &  & 27.92 & (0.61) & 6.00 &  & \framebox{\textbf{15.65}} & (0.56) & 8.96 &  & 16.50 & (0.46) & 6.07 \\
&DWD $q=1$  &  32.14 & (0.63) & 2.83 &  & 26.46 & (0.57) & 2.03 &  & 20.67 & (0.76) & 0.83 &  & 17.57 & (0.49) & 2.23 \\
&DWD $q=0.5$  &  \framebox{\textbf{31.62}} & (0.61) & 2.80 &  & \framebox{\textbf{26.42}} & (0.58) & 2.06 &  & 21.42 & (0.79) & 0.84 &  & 17.59 & (0.56) & 2.27 \\
&DWD $q=4$ & 31.63 & (0.61) & 3.05 &  & 26.42 & (0.57) & 2.08 &  & 20.26 & (0.76) & 0.91 &  & 17.15 & (0.50) & 2.28 \\
&DWD $q=8$ &  32.07 & (0.57) & 3.28 &  & 26.53 & (0.56) & 2.21 &  & 20.00 & (0.67) & 0.98 &  & 16.93 & (0.50) & 2.39 \\
   \bottomrule
\end{tabular}
 \end{minipage}}
\end{table}

\section{Discussion}
In this paper we have developed a new algorithm for solving the linear generalized DWD and the kernel generalized DWD. Compared with the current state-of-the-art algorithm for solving the linear DWD, our new algorithm is easier to understand, more general, and much more efficient. DWD equipped with the new algorithm can be computationally more efficient than the SVM. We have established the statistical learning theory of the kernel generalized DWD, showing that the kernel DWD and the kernel SVM are comparable in theory. Our theoretical analysis and algorithm do not suggest DWD with $q=1$ has any special merit compared to the other members in the generalized DWD family. Numerical examples further support our theoretical conclusions. DWD with $q=1$ is called the standard DWD purely due to the fact that it, not other generalized DWDs, can be solved by SOCP when the DWD idea was first proposed. Now with our new algorithm and theory, practitioners have the option to explore different DWD classifiers.

In the present paper we have considered the standard classification problem under the 0-1 loss. In many applications we may face the so-called non-standard classification problems. For example, observed data may be collected via biased sampling and/or we need to consider unequal costs
for different types of mis-classification. \cite{QiaoEtAl2010} introduced a weighted DWD to handle the non-standard classification problem, which follows the treatment of the non-standard SVM in \cite{LinEtAl2002}. \cite{QiaoEtAl2010} defined the weighted DWD as follows,
\begin{equation}
\begin{aligned}
\min_{\beta_0, \bs\beta} &\left[\sum_{i=1}^n w(y_i) \left(\dfrac{1}{r_i} + c\xi_i\right)\right], \text{ subject to } r_i = y_i(\beta_0 + \bs x_i^T\bs\beta)+\xi_i \ge 0 \text{ and } \bs\beta^T \bs\beta=1,
\end{aligned}
\label{eq:wDWD}
\end{equation}
which can be further generalized to the weighted kernel DWD:
\begin{equation}
\min_{\beta_0, \bs\alpha}\bs C_w(\beta_0, \bs\alpha) \equiv \min_{\beta_0, \bs\alpha} \left[\frac{1}{n} \sum_{i=1}^n w(y_i) V_q\left(y_i(\beta_0 + \bs K_i^T \bs\alpha)\right) + \lambda \bs \alpha^T \bs K \bs \alpha \right].
\label{eq:kerwtgenDWD}
\end{equation}
\cite{QiaoEtAl2010} gave the expressions for $w(y_i)$ for various non-standard classification problems. \cite{QiaoEtAl2010} solved the weighted DWD with $q=1$ \eqref{eq:wDWD} based on the second-order-cone programming. The MM procedure for Algorithm~\ref{alg:linear} and Algorithm~\ref{alg:kernel} can easily accommodate the weight factors $w(y_i)$'s to solve the weighted DWD and weighted kernel DWD. We have implemented the weighted DWD in the R package \texttt{kerndwd}.

\section*{Appendix: technical proofs}

\noindent {\bf \large Proof of Lemma~\ref{lm:DWD_loss}}

Write $v_i=y_i (\omega_0 + \bs{x}_i^T\bs{\omega})$ and $G(\eta_i)=1/{(v_i+\eta_i)^q}+c\eta_i$. The objective function of \eqref{eq:genDWD} can be written as $\sum_{i=1}^n G(\eta_i)$. We next minimize \eqref{eq:genDWD} over $\eta_i$ for every fixed $i$ by computing the first-order and the second-order derivatives of $G(\eta_i)$:
$$
\begin{aligned}
G'(\eta_i)&=-\dfrac{q}{(v_i+\eta_i)^{q+1}}+c=0\notag \Rightarrow v_i+\eta_i=\left(\dfrac{q}{c}\right)^{\frac{1}{q+1}},\notag\\
G''(\eta_i)&=\dfrac{q(q+1)}{(v_i+\eta_i)^{q+2}}>0.
\end{aligned}
$$
If $v_i>(\frac{q}{c})^{\frac{1}{q+1}}$, then $G'(\eta_i)>0$ for all $\eta_i \ge0$, and $\eta_i^\star=0$ is the minimizer. If $v_i\le(\frac{q}{c})^{\frac{1}{q+1}}$, then $\eta_i^\star=(\frac{q}{c})^{\frac{1}{q+1}}-v_i$ is the minimizer as $G'(\eta^\star)=0$ and $G''(\eta^\star)>0$.

By plugging in the minimizer $\eta_i^\star$ into $\sum_{i=1}^n G(\eta_i)$, we obtain
\begin{equation}
\begin{aligned}
\min_{\omega_0, \bs\omega} & \sum_{i=1}^n\tilde{V}_q\left(y_i (\omega_0 + \bs{x}_i^T\bs\omega)\right), \text{ subject to } \bs\omega^T \bs\omega = 1,
\end{aligned}
\label{eq:Vtilde}
\end{equation}
where
\begin{equation*}
\tilde{V}_q(v)=
\begin{cases}
      \left(\dfrac{q}{c}\right)^{-\frac{q}{q+1}}+c\left(\dfrac{q}{c}\right)^{\frac{1}{q+1}}-cv, \; &\text{ if } v \le  \left(\dfrac{q}{c}\right)^{\frac{1}{q+1}},\\
      \dfrac{1}{v^q}, \; &\text{ if } v >  \left(\dfrac{q}{c}\right)^{\frac{1}{q+1}}.
\end{cases}
\end{equation*}
We now simplify \eqref{eq:Vtilde}. Suppose $t=(\frac{q}{q+1})(\frac{q}{c})^{-\frac{1}{q+1}}$ and $t_1 = (\frac{1}{q+1})(\frac{q}{c})^{\frac{q}{q+1}}$. We define $V_q(u) = t_1 \cdot \tilde{V}_q(u/t)$ for each $q$,
$$
V_q(u)=
\begin{cases}
1-u, \; &\text{ if } u \le  \dfrac{q}{q+1},\\
      \dfrac{1}{u^q}\dfrac{q^q}{{(q+1)}^{q+1}}, \; &\text{ if } u > \dfrac{q}{q+1}.
    \end{cases}
$$
By setting $\beta_0 = t \cdot \omega_0$ and $\bs\beta = t\cdot \bs\omega$, we find that \eqref{eq:Vtilde} becomes
$$
\min_{\beta_0, \bs\beta} \sum_{i=1}^n V_q\left(y_i (\beta_0 + \bs x_i^T \bs\beta)\right), \text{ subject to } \bs\beta^T \bs\beta = t^2,
$$
which can be further transformed to \eqref{eq:DWDlossopt} with $\lambda$ and $t$ one-to-one correspondent.

\vspace{0.1in}
\noindent {\bf \large Proof of Lemma~\ref{lm:Lps}}

We first prove \eqref{eq:Lip_condtn}. We observe that $0<V''_q(u)=\frac{1}{u^{q+2}}\frac{q^{q+1}}{(q+1)^q}<\frac{(q+1)^2}{q}$, for any $u>\frac{q}{q+1}$. Also $V_q'(u)$ is continuous on $[\frac{q}{q+1}, \infty)$ and differentiable on $(\frac{q}{q+1}, \infty)$.

If both $u_1$ and $u_2>\frac{q}{q+1}$, then the mean value theorem implies that there exists $u^{\star\star}>\frac{q}{q+1}$, such that,
\begin{equation}
\dfrac{|V_q'(u_1)-V_q'(u_2)|}{|u_1-u_2|}=|V_q''(u^{\star\star})|<\dfrac{(q+1)^2}{q}.
\label{eq:A35}
\end{equation}

If $u_1>\frac{q}{q+1}$ and $u_2\le\frac{q}{q+1}$, then $V_q'(u_2)=V_q'\left(\frac{q}{q+1}\right)=-1$. The mean value theorem implies that there exists $u^{\star\star}>\frac{q}{q+1}$ satisfying
\begin{equation}
\dfrac{|V_q'(u_1)-V_q'(u_2)|}{|u_1-u_2|}\le\dfrac{|V_q'(u_1)-V_q'(\frac{q}{q+1})|}{|u_1-\frac{q}{q+1}|}=|V_q''(u^{\star\star})|<\dfrac{(q+1)^2}{q}.
\label{eq:A36}
\end{equation}

If both $u_1$ and $u_2\le\frac{q}{q+1}$, $V_q'(u_1)=V_q'(u_2)=-1$. It is trivial that
\begin{equation}
\dfrac{|V_q'(u_1)-V_q'(u_2)|}{|u_1-u_2|}=0<\dfrac{(q+1)^2}{q}.
\label{eq:A37}
\end{equation}
By \eqref{eq:A35}, \eqref{eq:A36}, and \eqref{eq:A37}, we prove \eqref{eq:Lip_condtn}.

We now prove \eqref{eq:quad_condtn}. 
Let
$
\nu(a)\equiv\dfrac{(q+1)^2}{2q}a^2-V_q(a).
$
From \eqref{eq:Lip_condtn}, it is not hard to show that
$
\nu'(a) = \dfrac{(q+1)^2}{q}a-V_q'(a)
$
is strictly increasing. Therefore $\nu(a)$ is a strictly convex function, and its first-order condition,
$
\nu(t) > \nu(\tilde{t}) + \nu'(\tilde{t})(t-\tilde{t}),
$
verifies \eqref{eq:quad_condtn} directly.

\vspace{0.1in}
\noindent {\bf \large Proof of Lemma~\ref{lm:Fisher}}

Given that $\eta(\bs x) = P(Y=1|\bs X=\bs x)$, we have that $E_{\bs XY} \left[V_q(Yf(\bs X))\right] \equiv E_{\bs X} \zeta(f(\bs X))$:
$$
\begin{aligned}
\zeta(f(\bs x)) &\equiv \eta(\bs x)V_q(f(\bs x)) + [1-\eta(\bs x)]V_q(-f(\bs x))\\
&=\begin{cases}
\eta(\bs x) \dfrac{1}{f(\bs x)^q} \dfrac{q^q}{(q+1)^{q+1}} + [1-\eta(\bs x)][1+f(\bs x)], &\text{ if } f(\bs x)>\dfrac{q}{q+1}, \\
\eta(\bs x)[1-f(\bs x)] + [1-\eta(\bs x)][1+f(\bs x)], &\text{ if } -\dfrac{q}{q+1} \le f(\bs x) \le \dfrac{q}{q+1}, \\
\eta(\bs x)[1-f(\bs x)] + [1-\eta(\bs x)]\dfrac{1}{[-f(\bs x)]^q}\dfrac{q^q}{(q+1)^{q+1}}, &\text{ if } f(\bs x)<-\dfrac{q}{q+1}.
\end{cases}
\end{aligned}
$$
For each given $\bs x$, we take both $f(\bs x)$ and $\eta(\bs x)$ as scalars and hereby write them as $f$ and $\eta$ respectively. We then take $\zeta(f)=\zeta(f(\bs x))$ as a function of $f$ and compute the derivative with respect to $f$:
$$
\dfrac{\partial \zeta(f)}{\partial f} =
\begin{cases}
-\eta \dfrac{1}{f^{q+1}} \dfrac{q^{q+1}}{(q+1)^{q+1}} + 1-\eta, &\text{ if }  f>\dfrac{q}{q+1}, \\
1-2\eta, &\text{ if }  -\dfrac{q}{q+1} \le f \le \dfrac{q}{q+1}, \\
-\eta + (1-\eta)\dfrac{1}{(-f)^{q+1}}\dfrac{q^{q+1}}{(q+1)^{q+1}}, &\text{ if }  f<-\dfrac{q}{q+1}.
\end{cases}
$$
We see that (1) when $\eta > 0.5$, $\partial \zeta(f)/\partial f = 0$ only when $f = \tilde{f} \equiv \frac{q}{q+1}\left(\frac{\eta}{1-\eta}\right)^{\frac{1}{q+1}}$, and (2) when $\eta < 0.5$, $\partial \zeta(f)/\partial f = 0$ only when $f = \tilde{f} \equiv -\frac{q}{q+1}\left(\frac{1-\eta}{\eta}\right)^{\frac{1}{q+1}}$. 
For these two cases, we also observe that 
\begin{equation}
\begin{cases}
\partial \zeta(f)/\partial f < 0, \text{ if } f < \tilde{f}, \\
\partial \zeta(f)/\partial f > 0, \text{ if } f > \tilde{f},
\end{cases}
\label{eq:deriv_f}
\end{equation}
which follows that $\tilde{f}$ is the minimizer of $\zeta(f)$. 

\vspace{0.1in}
\noindent{\bf \large Proof of Lemma~\ref{lm:err_bd}}

As $\tilde{f}(\bs x)$ was defined in \eqref{eq:Fisher_min}, we see that for each $\bs x$,
\begin{equation*}
\begin{aligned}
\zeta\left(\tilde{f}(\bs x)\right) &\equiv \eta(\bs x)V_q\left(\tilde{f}(\bs x)\right)+[1-\eta(\bs x)] V_q\left(-\tilde{f}(\bs x)\right)\\
&=
\begin{cases}
\eta(\bs x) + [1-\eta(\bs x)]^{\frac{1}{q+1}}\eta(\bs x)^{\frac{q}{q+1}}, & \text{ if } \eta(\bs x) \le 1/2,\\
1 - \eta(\bs x) + \eta(\bs x)^{\frac{1}{q+1}}[1-\eta(\bs x)]^{\frac{q}{q+1}}, & \text{ if } \eta(\bs x) > 1/2, 
\end{cases}\\
&= \dfrac{1}{2}\bigg(1-|2\eta(\bs x)-1|\bigg) + \dfrac{1}{2}\bigg(1+|2\eta(\bs x)-1|\bigg)^{\frac{1}{q+1}}\bigg(1-|2\eta(\bs x)-1|\bigg)^{\frac{q}{q+1}}.
\end{aligned}
\end{equation*}
For $a\in [0, 1]$, we define $\gamma(a)$ and compute its first-order derivative as follows,
$$
\begin{aligned}
\gamma(a) &\equiv 1 - \dfrac{1}{2}(1-a) - \dfrac{1}{2}(1+a)^{\frac{1}{q+1}}(1-a)^{\frac{q}{q+1}} - \dfrac{q}{q+1}a, \\
\gamma'(a) &= \dfrac{1}{2} - \dfrac{1}{2(q+1)}\left(\dfrac{1-a}{1+a}\right)^{\frac{q}{q+1}} + \dfrac{q}{2(q+1)}\left(\dfrac{1+a}{1-a}\right)^{\frac{1}{q+1}} - \dfrac{q}{q+1}\\
&=\left[\dfrac{1}{2(q+1)} - \dfrac{1}{2(q+1)}\left(\dfrac{1-a}{1+a}\right)^{\frac{q}{q+1}}\right] + \left[\dfrac{q}{2(q+1)} + \dfrac{q}{2(q+1)}\left(\dfrac{1+a}{1-a}\right)^{\frac{1}{q+1}} - \dfrac{q}{q+1}\right] \ge 0.
\end{aligned}
$$
Hence for each $a\in [0,1]$, $\gamma(a) \ge \gamma(0) = 0$. For each $\bs x$, let $a=|2\eta(\bs x)-1|$ and we see that
$$
1-\zeta\left(\tilde{f}(\bs x)\right) \ge \dfrac{q}{q+1}|2\eta(\bs x)-1|.
$$
By
$R(f) = E_{\bs XY}[Y \neq \mathrm{sign}(f(\bs X)] =E_{\{\bs X: f(\bs X)\ge 0\}}[1-\eta(\bs X)] + E_{\{\bs X: f(\bs X)\le 0\}}\eta(\bs X),
$
we obtain 
\begin{equation}
\begin{aligned}
R(\hat{f}_n) - R(f^\star) &=E_{\{\bs X: \hat{f}_n(\bs X) \ge 0,\ f^{\star}(\bs X) <0\}}[1-2\eta(\bs X)] + 
E_{\{\bs X: \hat{f}_n(\bs X) \le 0,\ f^\star(\bs X) >0\}}[2\eta(\bs X)-1]\\
&\le E_{\{\bs X: \hat{f}_n(\bs X)f^\star(\bs X)\le 0\}}|2\eta(\bs X)-1|\\
&\le \dfrac{q+1}{q}	E_{\{\bs X: \hat{f}_n(\bs X)f^\star(\bs X)\le 0\}} \left[1 - \zeta\left(\tilde{f}(\bs X)\right)\right].
\end{aligned}
\label{eq:Rf}
\end{equation}
Since $f^\star(\bs X)$ and $\tilde{f}(\bs X)$ share the same sign, $\hat{f}_n(\bs X)f^\star(\bs X) \le 0$ implies that $\hat{f}_n(\bs X)\tilde{f}(\bs X) \le 0$. When $\hat{f}_n(\bs X)\tilde{f}(\bs X) \le 0$, 0 is between $\hat{f}_n(\bs X)$ and $\tilde{f}(\bs X)$, and thus \eqref{eq:deriv_f} indicates that $\zeta(\tilde{f}(\bs X)) \le \zeta(0) = 1 \le \zeta(\hat{f}_n(\bs X))$.
From \eqref{eq:Rf}, we conclude that
\begin{equation*}
\begin{aligned}
R(\hat{f}_n) - R(f^\star) &\le \dfrac{q+1}{q}	E_{\{\bs X: \hat{f}_n(\bs X)f^\star(\bs X)\le 0\}} \left[\zeta\left(\hat{f}_n(\bs X)\right) - \zeta\left(\tilde{f}(\bs X)\right) \right]\\
&\le \dfrac{q+1}{q}	E_{\bs X} \left[\zeta\left(\hat{f}_n(\bs X)\right) - \zeta\left(\tilde{f}(\bs X)\right) \right]\\
&=\dfrac{q+1}{q}	E_{\bs XY} \left[V_q\left(Y\hat{f}_n(\bs X)\right) - V_q\left(Y\tilde{f}(\bs X)\right) \right]\\
&=\dfrac{q+1}{q}(\varepsilon_A + \varepsilon_E).
\end{aligned}
\end{equation*}

\vspace{0.1in}
\noindent {\bf \large Proof of Theorem~\ref{lm:thm}}

{\bf Part} (1).  We first show that when $\mathcal{H}_K$ is induced by a universal kernel, the approximation error $\varepsilon_A = 0$. 
By definition, we need to show that for any $\epsilon>0$, there exists $f_{\epsilon} \in \mathcal{H}_K$ such that
\begin{equation}
\bigg|E_{\bs XY}V_q\left(Yf_{\epsilon}(\bs X)\right) - E_{\bs XY}V_q\left(Y\tilde{f}(\bs X)\right)\bigg| < \epsilon.
\label{eq:proof1}
\end{equation}

We first use truncation to consider a truncated version of $\tilde f$. For any given $\delta \in (0, 0.5)$, we define
$$
f_\delta(\bs X) = 
\begin{cases}
\frac{q}{q+1}\left(\frac{1-\delta}{\delta} \right)^\frac{1}{q+1}, & \text{ if } \eta(\bs X) > 1-\delta,  \\
\tilde{f}(\bs X), & \text{ if } -\delta \le \eta(\bs X) \le 1-\delta,\\
-\frac{q}{q+1}\left(\frac{\delta}{1-\delta} \right)^\frac{1}{q+1}, & \text{ if } \eta(\bs X) < \delta.
\end{cases}
$$
We have that
$$
0 \le E_{\bs XY} V_q \left(Y f_\delta(\bs X) \right) - E_{\bs XY}V_q \left(Y\tilde{f}(\bs X)\right) = \kappa_+ + \kappa_-,
$$
where 
\begin{equation*}
\begin{aligned}
\kappa_+ = &E_{\bs X: \eta(\bs X)>1-\delta}\left[ \eta(\bs X) V_q(f_\delta(\bs X))+(1-\eta(\bs X)) V_q(-f_\delta(\bs X)) \right]\\
&-E_{\bs X: \eta(\bs X)>1-\delta}\left[ \eta(\bs X) V_q \left(\tilde{f}(\bs X)\right)   +  (1-\eta(\bs X)) V_q\left(-\tilde{f}(\bs X)\right) \right],\\
\kappa_- = &E_{\bs X: \eta(\bs X)<\delta}\left[ \eta(\bs X) V_q(f_\delta(\bs X)) + (1-\eta(\bs X)) V_q(-f_\delta(\bs X)) \right]\\
&-E_{\bs X: \eta(\bs X)<\delta} \left[ \eta(\bs X) V_q \left(\tilde{f}(\bs X)\right)   +  (1-\eta(\bs X)) V_q\left(-\tilde{f}(\bs X)\right) \right].
\end{aligned}
\end{equation*}
Since $V_q(f_\delta(\bs X)) < V_q(-f_\delta(\bs X))$ when $\eta(\bs X)>1-\delta$, 
\begin{equation*}
\begin{aligned}
\kappa_+ < &E_{\bs X: \eta(\bs X)>1-\delta}\left[ (1-\delta) V_q(f_\delta(\bs X)) + \delta V_q(-f_\delta(\bs X)) \right]\\
&-E_{\bs X: \eta(\bs X)>1-\delta}\left[ \eta(\bs X) V_q\left(\tilde{f}(\bs X)\right)   +  (1-\eta(\bs X)) V_q\left(-\tilde{f}(\bs X)\right) \right]\\
= & \left[\delta + (1-\delta)^{\frac{1}{q+1}}\delta^\frac{q}{q+1} \right] - E_{\bs X: \eta(\bs X)>1-\delta} \left[1-\eta(\bs X) + \eta(\bs X)^{\frac{1}{q+1}}(1-\eta(\bs X))^{\frac{q}{q+1}} \right].
\end{aligned}
\end{equation*}
We notice that $(1-a) + a^{\frac{1}{q+1}}(1-a)^\frac{q}{q+1}$ is a continuous function in terms of $a \in (0,1)$. Since $\eta(\bs X)>1-\delta$ implies that $|\eta(\bs X) - (1 - \delta)| < \delta$, we conclude that for any given $\epsilon > 0$, there exists a sufficiently small $\delta$ such that $\kappa_+ < \epsilon/6$. We can also obtain $\kappa_- < \epsilon/6$ in the same spirit. Therefore,
\begin{equation}
0 \le E_{\bs XY} V_q \left(Y f_\delta(\bs X) \right) - E_{\bs XY}V_q \left(Y\tilde{f}(\bs X)\right) \le \kappa_+ + \kappa_- < \epsilon/3.
\label{eq:eA_p1}
\end{equation}

By Lusin's Theorem, there exists a continuous function $\varrho(\bs X)$ such that $P(\varrho(\bs X)\neq f_{\delta}(\bs X)) \le \epsilon(q+1)/(6q)$. Notice that $\sup_{\bs X}|f_\delta(\bs X)| \le q/(q+1)$.  Define 
$$
\tau(\bs X) = 
\begin{cases}
\varrho(\bs X), & \text{ if } |\varrho(\bs X)| \le \dfrac{q}{q+1},\\
\dfrac{q}{q+1}\cdot\dfrac{\varrho(\bs X)}{|\varrho(\bs X)|}, & \text{ if } |\varrho(\bs X)| > \dfrac{q}{q+1},\\
\end{cases} 
$$
then $P(\tau(\bs X)\neq f_{\delta}(\bs X)) \le \epsilon(q+1)/(6q)$ as well. Hence
\begin{equation}
\begin{aligned}
\bigg|E_{\bs XY} V_q \left(Y f_\delta(\bs X) \right) - E_{\bs XY} V_q \left(Y \tau(\bs X) \right)\bigg|
&\le& E_{\bs X}|f_{\delta}(\bs X) - \tau(\bs X)| \nonumber \\
&=& E_{\{\bs X: \tau(\bs X)\neq f_{\delta}(\bs X)\}}|f_{\delta}(\bs X) - \tau(\bs X)| \nonumber \\
&\le& \dfrac{2q}{q+1} \cdot \dfrac{\epsilon(q+1)}{6q} =\epsilon/3,
\label{eq:eA_p2}
\end{aligned}
\end{equation}
where the first inequality comes from the fact that $V_q(u)$ is Lipschitz continuous, i.e.,
$$
|V_q(u_1) - V_q(u_2)| \le |u_1 - u_2|, \ \forall u_1, u_2 \in \mathbb{R}.
$$
Notice that $\tau(\bs X)$ is also continuous. The definition of the universal kernel implies the existence of a function $f_{\epsilon} \in \mathcal{H}_K$ such that
\begin{equation}
\begin{aligned}
&\bigg|E_{\bs XY} V_q \left(Y f_{\epsilon}(\bs X) \right) - E_{\bs XY} V_q \left(Y \tau(\bs X) \right)\bigg| < \sup_{\bs X}|f_{\epsilon}(\bs X) - \tau(\bs X)| < \epsilon/3.
\label{eq:eA_p3}
\end{aligned}
\end{equation}
By combining \eqref{eq:eA_p1}, \eqref{eq:eA_p2}, and \eqref{eq:eA_p3} we obtain \eqref{eq:proof1}.

{\bf Part} (2). In this part we bound the estimation error $\varepsilon_E(\hat{f}_n)$. 
Note that RKHS has the following reproducing property \citep{Wahba1990, HastieEtAl2009}:
\begin{equation}
\begin{aligned}
&\langle K(\bs x_i, \bs x), f(\bs x)\rangle_{\mathcal{H}_K} = f(\bs x_i),\\
&\langle K(\bs x_i, \bs x), K(\bs x_j, \bs x)\rangle_{\mathcal{H}_K} = K(\bs x_i, \bs x_j).
\end{aligned}
\end{equation}
Fix any $\epsilon>0$. By the KKT condition of \eqref{eq:ker_DWD} and the representor theorem, we have 
\begin{equation}
\dfrac{1}{n}\sum_{i=1}^n V'_q\left(y_i  \hat{f}_n(\bs x_i) \right) y_i K(\bs x_i, \bs x) + 2\lambda_n \hat{f}_n(\bs x)=0.
\label{eq:ker_dwd_min}
\end{equation}
We define $\hat{f}^{[k]}$ as the solution of \eqref{eq:ker_DWD} when the $k$th observation is excluded from the training data, i.e.,
\begin{equation}
\hat{f}^{[k]} = \argmin_{f \in {\mathcal{H}}_K} \left[ \dfrac{1}{n}\sum_{i=1, i \neq k}^n V_q \left( y_i (f(\bs x_i) \right) + \lambda_n||f||^2_{{\mathcal{H}}_K} \right].
\label{eq:ker_DWD_delete_k}
\end{equation}
By the definition of $\hat{f}^{[k]}$ and the convexity of $V_q$, we have
\begin{equation*}
\begin{aligned}
0\le&  \dfrac{1}{n}\sum_{i=1, i\neq k}^{n}V_q\left(y_i \hat{f}_n(\bs x_i) \right) + \lambda_n ||\hat{f}_n||^2_{\mathcal{H}_K}-\dfrac{1}{n}\sum_{i=1, i\neq k}^{n}V_q\left(y_i \hat{f}^{[k]}(\bs x_i) \right) - \lambda_n ||\hat{f}^{[k]}||^2_{\mathcal{H}_K}\\
\le &  - \dfrac{1}{n}\sum_{i=1, i\neq k}^{n}V'_q\left(y_i  \hat{f}_n(\bs x_i) \right) y_i\left(\hat{f}^{[k]} (\bs x_i) - \hat{f}_n(\bs x_i)\right) + \lambda_n ||\hat{f}_n||^2_{\mathcal{H}_K}
-\lambda_n ||\hat{f}^{[k]}||^2_{\mathcal{H}_K}.
\end{aligned}
\end{equation*}
By the reproducing property, we further have
\begin{equation*}
\begin{aligned}
0\le & - \dfrac{1}{n}\sum_{i=1, i\neq k}^{n}V'_q\left(y_i  \hat{f}_n(\bs x_i) \right) y_i\left\langle  K(\bs x_i, \bs x), \hat{f}^{[k]} (\bs x) - \hat{f}_n(\bs x)\right\rangle_{\mathcal{H}_K} + \lambda_n ||\hat{f}_n||^2_{\mathcal{H}_K}
-\lambda_n ||\hat{f}^{[k]}||^2_{\mathcal{H}_K}\\
= & - \dfrac{1}{n}\sum_{i=1, i\neq k}^{n}V'_q\left(y_i  \hat{f}_n(\bs x_i) \right) y_i\left\langle  K(\bs x_i, \bs x), \hat{f}^{[k]} (\bs x) - \hat{f}_n(\bs x)\right\rangle_{\mathcal{H}_K} \\
& - 2 \lambda_n \left\langle \hat{f}_n(\bs x), \hat{f}^{[k]}(\bs x)-\hat{f}_n(\bs x) \right\rangle_{\mathcal{H}_K} - \lambda_n||\hat{f}^{[k]}- \hat{f}_n||^2_{\mathcal{H}_K}\\
= & \dfrac{1}{n} V'_q\left(y_k \hat{f}_n(\bs x_k) \right) y_k \left\langle  K(\bs x_k, \bs x), \hat{f}^{[k]} (\bs x) - \hat{f}_n(\bs x)\right\rangle_{\mathcal{H}_K} - \lambda_n||\hat{f}^{[k]}- \hat{f}_n||^2_{\mathcal{H}_K},
\end{aligned}
\end{equation*}
where the equality in the end holds by \eqref{eq:ker_dwd_min}. Thus, by Cauchy-Schwartz inequality,
$$
\begin{aligned}
&n\lambda_n||\hat{f}^{[k]}- \hat{f}_n||^2_{\mathcal{H}_K}
\le V'_q\left(y_k \hat{f}_n(\bs x_k) \right) y_k \left\langle  K(\bs x_k, \bs x), \hat{f}^{[k]} (\bs x) - \hat{f}_n(\bs x)\right\rangle_{\mathcal{H}_K}\\
\le& \left|V'_q\left(y_k \hat{f}_n(\bs x_k) \right) \right| ||K(\bs x_k, \bs x)||_{\mathcal{H}_K}||\hat{f}^{[k]}- \hat{f}_n||_{\mathcal{H}_K}
\le \sqrt{K(\bs x_k, \bs x_k)} \cdot || \hat{f}^{[k]}- \hat{f}_n||_{\mathcal{H}_K},
\end{aligned}
$$
which implies 
$$
||\hat{f}^{[k]}- \hat{f}_n||_{\mathcal{H}_K} \le \dfrac{\sqrt{B}}{n\lambda_n},
$$
where $B=\sup_{\bs x} K(\bs x, \bs x)$. 
By the reproducing property, we have
$$
\begin{aligned}
|\hat{f}^{[k]}(\bs x_k) - \hat{f}_n(\bs x_k)|^2&=\left(\langle K(\bs x_i, \bs x_k), \hat{f}^{[k]}(\bs x_i) - \hat{f}_n(\bs x_i)\rangle_{\mathcal{H}_K}\right)^2 \\
&\le  K(\bs x_k, \bs x_k)  ||\hat{f}^{[k]}- \hat{f}_n||^2_{\mathcal{H}_K} \le B \left(\dfrac{\sqrt{B}}{n\lambda_n}\right)^2.
\end{aligned}
$$
By the Lipschitz continuity of the DWD loss, we obtain that for each $k=1,\ldots, n$,
$$
\begin{aligned}
&V_q\left(y_k\hat{f}^{[k]}(\bs x_k) \right) - V_q\left(y_k\hat{f}_n(\bs x_k) \right)
\le & |\hat{f}^{[k]}(\bs x_k) - \hat{f}_n(\bs x_k)|
\le \dfrac{B}{n\lambda_n},
\end{aligned}
$$
and therefore,
\begin{equation}
\dfrac{1}{n}\sum_{k=1}^n V_q\left(y_k\hat{f}^{[k]}(\bs x_k) \right) \le  \dfrac{1}{n}\sum_{k=1}^n V_q\left(y_k\hat{f}_n(\bs x_k) \right) +  \dfrac{B}{n\lambda_n}.
\label{eq:proof_eq100}
\end{equation}
Let $f^*_{\epsilon} \in \mathcal{H}_{K}$ such that  
\begin{equation}
E_{\bs XY} V_q\left(Yf^*_{\epsilon}(\bs X) \right) \le \inf_{f\in\mathcal{H}_K}E_{\bs XY} V_q\left(Yf(\bs X) \right)+\epsilon/3.
\label{eq:proof_eq101}
\end{equation}
By definition of $\hat{f}_n$, we have
\begin{equation}
\dfrac{1}{n}\sum_{k=1}^n V_q\left(y_k\hat{f}_n(\bs x_k) \right)+\lambda_n  ||\hat{f}_n||^2_{\mathcal{H}_K} \le 
\dfrac{1}{n}\sum_{k=1}^n V_q\left(y_k f^*_{\epsilon}(\bs x_k) \right)+\lambda_n || f^*_{\epsilon}||^2_{\mathcal{H}_K}. 
\label{eq:proof_eq102}
\end{equation}
Since each data point in $\bs T_n = \{(\bs x_k, y_k)\}_{k=1}^n$ is drawn from the same distribution, we have
\begin{equation}
\begin{aligned}
E_{\bs T_n} \left[\dfrac{1}{n}\sum_{k=1}^n V_q\left(y_k\hat{f}^{[k]}(\bs x_k) \right) \right] = \dfrac{1}{n}\sum_{k=1}^n E_{\bs T_n}V_q\left(y_k\hat{f}^{[k]}(\bs x_k) \right) = E_{\bs T_{n-1}}E_{\bs XY} V_q\left(Y\hat{f}_{n-1}(\bs X) \right). \quad \label{eq:proof_eq103}
\end{aligned}
\end{equation}
By combining \eqref{eq:proof_eq100}--\eqref{eq:proof_eq103} we have 
\begin{equation}
E_{\bs T_{n-1}}E_{\bs XY} V_q\left(Y\hat{f}_{n-1}(\bs X) \right)  \le \inf_{f\in\mathcal{H}_K}E_{\bs XY} V_q\left(Yf(\bs X) \right)
+\lambda_n ||f^*_{\epsilon}||^2_{\mathcal{H}_K} +  \dfrac{B}{n\lambda_n}+\frac{\epsilon}{3}.
\label{eq:proof_eq105}
\end{equation}
By the choice of $\lambda_n$, we see that there exits $N_{\epsilon}$ such that when $n>N_{\epsilon}$ we have $\lambda_n <\epsilon/(3||f^*_{\epsilon}||^2_{\mathcal{H}_K})$, $n\lambda_n>3B/\epsilon$, and hence
$$
E_{\bs T_{n-1}}\left[E_{\bs XY} V_q\left(Y\hat{f}_{n-1}(\bs X) \right)\right] \le  \inf_{f\in\mathcal{H}_K}E_{\bs XY} V_q\left(Yf(\bs X) \right)+\epsilon.
$$
Because $\epsilon$ is arbitrary and $
E_{\bs T_{n-1}}[E_{\bs XY} V_q(Y\hat{f}_{n-1}(\bs X) )] \ge  \inf_{f\in\mathcal{H}_K}E_{\bs XY} V_q\left(Yf(\bs X) \right)
$, we have $\lim_{n\to\infty} E_{\bs T_{n-1}}[E_{\bs XY} V_q(Y\hat{f}_{n-1}(\bs X) )] = \inf_{f\in\mathcal{H}_K}E_{\bs XY} V_q\left(Yf(\bs X) \right)$,
which equivalently indicates that
$
\lim_{n \to\infty} E_{\bs T_{n}}\varepsilon_E(\hat{f}_n) = 0.
$
Since $\varepsilon_E(\hat{f}_n) \ge 0$, then by Markov inequality, we prove part (2).



\begin{thebibliography}{31}
\newcommand{\enquote}[1]{``#1''}
\expandafter\ifx\csname natexlab\endcsname\relax\def\natexlab#1{#1}\fi
\providecommand{\natexlab}[1]{#1}

\bibitem[{Ahn et~al.(2007) Ahn, Marron, Muller, and Chi}]{AhnEtAl2007}
Ahn, J., Marron, J.S., Muller, K., and Chi, Y. (2007), \enquote{{The high-dimension, low-sample-size geometric representation holds under mild conditions},} \textit{Biometrika}, 94(3), 760--766.

\bibitem[{Ahn and Marron(2010)}]{AhnMarron2010}
Ahn, J. and Marron, J.S. (2010), \enquote{{The maximal data-piling direction for discrimination},} \textit{Biometrika}, 97(1), 254--259.

\bibitem[{Aizerman et~al.(1964) Aizerman, Braverman, and Rozoner}]{AizermanEtAl1964}
Aizerman, A., Braverman, E., and Rozoner, L. (1964), \enquote{{Theoretical foundations of the potential function method in pattern recognition learning},} \textit{Automation and remote control}, 25, 821--837.

\bibitem[{Alizadeh and Goldfarb(2004)}]{AlizadehGoldfarb2004}
Alizadeh, F. and Goldfarb, D. (2004), \enquote{{Second-order cone programming},} \textit{Mathematical Programming, Series B}, 95(1), 3--51.

\bibitem[{Anthony and Bartlett(1999)}]{AnthonyBartlett1999}
Anthony, M. and Bartlett, P. (1999), \textit{Neural Network Learning: Theoretical Foundations.}, Cambridge University Press, Cambridge.

\bibitem[{Bartlett and Shawe-Taylor(1999)}]{BartlettShaweTaylor1999}
Bartlett, P. and Shawe-Taylor, J. (1999), \enquote{Generalization performance of support vector machines and other pattern classifiers}, \textit{Advances in Kernel Methods--Support Vector Learning}, 43--54.


\bibitem[{Boyd and Vandenberghe(2004)}]{BoydVandenberghe2004}
Boyd, S. and Vandenberghe, L. (2004), \textit{Convex Optimization}, Cambridge University Press, Cambridge.

\bibitem[{Breiman(2001)}]{Breiman2001}
Breiman, L. (2001), \enquote{Random forests,} \textit{Machine Learning}, 45(1), 5--32.


\bibitem[{De Leeuw and Heiser(1977)}]{DeLeeuwHeiser1977}
De Leeuw, J. and Heiser, W. (1977), \enquote{Convergence of correction matrix algorithms for multidimensional scaling}, 735--752.

\bibitem[{Fern{\'a}ndez-Delgado et~al.(2014) Fern{\'a}ndez-Delgado, Cernadas, Barro, and Amorim}]{FernandezEtAl2014}
Fern{\'a}ndez-Delgado, M., Cernadas, E., Barro, S., and Amorim, D. (2014), \enquote{{Do we need hundreds of classifiers to solve real world classification problems}?} \textit{The Journal of Machine Learning Research}, 15, 3133--3181.

\bibitem[{Freund and Schapire(1997)}]{FreundSchapire1997}
Freund, Y. and Schapire, R. (1997),
\enquote{A decision-theoretic generalization of on-line learning and an application to boosting,} \textit{Journal of Computer and System Sciences}, 55(1), 119--139.

\bibitem[{Friedman et~al.(2007) Friedman, Hastie, H{\"o}fling, and Tibshirani}]{FriedmanEtAl2007}
Friedman, J., Hastie, T., H{\"o}fling, H., and Tibshirani, R. (2007), \enquote{{Pathwise coordinate optimization},} \textit{The Annals of Applied Statistics}, 1(2), 302--332.

\bibitem[{Girosi et~al.(1995) Girosi, Jones, and Poggio}]{GirosiEtAl1995}
Girosi, F., Jones, M., and Poggio, T. (1995), \enquote{{Regularization theory and neural networks architectures},} \textit{Neural Computation}, 7(2), 219--269.

\bibitem[{Hall et~al.(2005) Hall, Marron, and Neeman}]{HallEtAl2005}
Hall, P., Marron, J.S., and Neeman, A. (2005), \enquote{{Geometric representation of high dimensions, low sample size data},} \textit{Journal of the Royal Statistical Society, Series B}, 67(3), 427--444.

\bibitem[{Hastie et~al.(2009) Hastie, Tibshirani, and Friedman}]{HastieEtAl2009}
Hastie, T., Tibshirani, R., and Friedman, J. (2009), \textit{The Elements of Statistical Learning: Prediction, Inference, and Data Mining}, 2nd edition, Springer-Verlag, New York.


\bibitem[{Huang et~al.(2013)}]{HuangEtAl2013}
Huang, H., Liu, Y., Du. Y., Perou, C., Hayes, D., Todd, M., and Marron, J.S. (2013), \enquote{{Multiclass distance-weighted discrimination},} \textit{Journal of Computational and Graphical Statistics}, 22(4), 953--969.

\bibitem[{Huang et~al.(2012)}]{HuangEtAl2012}
Huang, H., Lu, X., Liu, Y., Haaland, P., and Marron, J.S. (2012), \enquote{{R/DWD: distance-weighted discrimination for classification, visualization and batch adjustment},}
\textit{Bioinformatics}, 28(8), 1182--1183.

\bibitem[{Hunter and Lange(2004)}]{HunterLange2004}
Hunter, D. and Lange, K. (2004), \enquote{{A tutorial on MM algorithms},} \textit{The American Statistician}, 58(1), 30--37.

\bibitem[{Hunter and Li(2005)}]{HunterLi2005}
Hunter, D. and Li, R. (2005), \enquote{{Variable selection using MM algorithms},} \textit{The Annals of Statistics}, 33(4), 1617--1642.

\bibitem[{Jaakkola and Haussler(1999)}]{JaakkolaHaussler1999}
Jaakkola, T. and Haussler, D. (1999), \enquote{{Probabilistic kernel regression models},} \textit{Proceedings of the 1999 Conference on AI and Statistics}, 126, 00--04.

\bibitem[{Karatzoglou et~al.(2004)Karatzoglou, Smola, Hornik, and Zeileis}]{KaratzoglouEtAl2004}
Karatzoglou, A., Smola, A., Hornik, K., and Zeileis, A.(2004), \enquote{{kernlab -- An S4 Package for Kernel Methods in R},} \textit{Journal of Statistical Software}, 11(9), 1--20.

\bibitem[{Lange et~al.(2000)Lange, Hunter, and Yang}]{LangeEtAl2000}
Lange, K., Hunter, D., and Yang, I. (2000), \enquote{{Optimization transfer using surrogate objective functions},} \textit{Journal of Computational and Graphical Statistics}, 9(1), 1--20.

\bibitem[{Lange and Zhou(2014)}]{LangeZhou2014}
Lange, K. and Zhou, H. (2014), \enquote{{MM algorithms for geometric and signomial programming},}
\textit{Mathematical Programming}, 143(1-2), 339--356.

\bibitem[{Lichman(2013)}]{Lichman2013}
Lichman, M. (2013), \enquote{{UCI Machine Learning Repository},}
\url{http://archive.ics.uci.edu/ml}, Irvine, CA: University of California, School of Information and Computer Science.

\bibitem[{Lin et~al.(2002)}]{LinEtAl2002}
Lin, Y., Lee, Y., and Wahba, G. (2002), \enquote{{Support vector machines for classification in nonstandard situations},}\textit{Machine Learning}, 46, 191--202.

\bibitem[{Lin(2002)}]{Lin2002}
Lin, Y. (2002), \enquote{{Support vector machines and the Bayes rule in classification},}
\textit{Data Mining and Knowledge Discovery}, 6(3), 259--275.

\bibitem[{Lin(2004)}]{Lin2004}
Lin, Y. (2004), \enquote{{A note on margin-based loss functions in classification},}
\textit{Statistics \& Probability Letters}, 68(1), 73--82.


\bibitem[{Liu et~al.(2011)}]{LiuEtAl2011}
Liu, Y., Zhang, H., and Wu, Y. (2011), \enquote{{Hard or soft classification? Large-margin unified machines},}
\textit{Journal of American Statistical Association}, 106(493), 166--177.

\bibitem[{Marron et~al.(2007)Marron, Todd, and Ahn}]{MarronEtAl2007}
Marron, J.S., Todd, M., and Ahn, J. (2007), \enquote{{Distance weighted discrimination},} \textit{Journal of American Statistical Association}, 102(480), 1267--1271.

\bibitem[{Marron(2013)}]{Marron2013}
Marron, J.S. (2013), \enquote{{Smoothing, functional data analysis, and distance weighted discrimination software},}
\url{http://www.unc.edu/~marron/marron_software.html}.

\bibitem[{Marron(2015)}]{Marron2015}
Marron, J.S. (2015), \enquote{{Distance-weighted discrimination},}
\textit{Wiley Interdisciplinary Reviews: Computational Statistics}, 7(2), 109--114.

\bibitem[{Micchelli et~al.(2006)Micchelli, Xu, and Zhang}]{MicchelliEtAl2006}
Micchelli, C., Xu, Y., and Zhang, H. (2006), \enquote{{Universal kernels},} \textit{Journal of Machine Learning Research}, 7, 2651--2667.



\bibitem[{Qiao et~al.(2010)Qiao, Zhang, Liu, Todd, and Marron}]{QiaoEtAl2010}
Qiao, X., Zhang, H., Liu, Y., Todd, M., Marron, J.S. (2010), \enquote{{Weighted distance weighted discrimination and its asymptotic properties},} \textit{Journal of American Statistical Association}, 105(489), 401--414.

\bibitem[{Qiao and Zhang (2015a)}]{QiaoZhang2015a}
Qiao, X. and Zhang, L. (2015a), \enquote{{Distance-weighted support vector machine},} \textit{Statistics and Its Interface}, 8(3), 331--345.

\bibitem[{Qiao and Zhang (2015b)}]{QiaoZhang2015b}
Qiao, X. and Zhang, L. (2015b), \enquote{{Flexible high-dimensional classification machines and their asymptotic properties},} \textit{Journal of Machine Learning Research}, forthcoming.

%
%


\bibitem[{Shawe-Taylor and Cristianini(2000)}]{ShaweTaylorCristianini2000}
Shawe-Taylor, J. and Cristianini, N. (2000), \enquote{Margin distribution and soft margin}, \textit{Advances in Kernel Methods--Support Vector Learning}, 349--358.

\bibitem[{Steinwart(2001)}]{Steinwart2001}
Steinwart, I. (2001), \enquote{{On the influence of the kernel on the consistency of support vector machines},} \textit{Journal of Machine Learning Research}, 2, 67--93.

\bibitem[{T\"ut\"unc\"u et~al.(2003)}]{TutuncuEtAl2003}
T\"ut\"unc\"u R., Toh, K., Todd, M. (2003), \enquote{{Solving semidefinite-quadratic-linear programs using SDPT3},} \textit{Mathematical Programming}, 95(2), 189--217.



\bibitem[{Vapnik(1995)}]{Vapnik1995}
Vapnik, V. (1995), \textit{The Nature of Statistical Learning Theory}, Springer-Verlag, New York.

\bibitem[{Vapnik(1998)}]{Vapnik1998}
Vapnik, V. (1998), \textit{Statisitcal Learning Theory}, Wiley, New York.

\bibitem[{Wahba(1990)}]{Wahba1990}
Wahba, G. (1990), \textit{Spline Models for Observational Data}, 59, SIAM.

\bibitem[{Wahba et~al.(1994)}]{WahbaEtAl1994}
Wahba, G., Gu, C., Wang, Y., and Campbell, R. (1994), \enquote{{Soft classification, aka risk estimation, via penalized log likelihood and smoothing spline analysis of variance},} In \textit{Santa fe Institute Studies in the Sciences of Complexity-Proceeding Vol}, 20, Addison-Wesley Publishing CO, 331--331.

\bibitem[{Wahba(1999)}]{Wahba1999}
Wahba, G. (1999), \enquote{{Support vector machines, reproducing kernel Hilbert spaces and the randomized GACV},} \textit{Advances in Kernel Methods-Support Vector Learning}, 6, 69--87.
%

\bibitem[{Wang and Zou(2015)}]{WangZou2015}
Wang, B. and Zou, H. (2015), \enquote{{Sparse distance weighted discrimination},} \textit{Journal of Computational and Graphical Statistics}, forthcoming.

\bibitem[{Wu and Lange(2008)}]{WuLange2008}
Wu, T.T. and Lange, K. (2008), \enquote{{Coordinate descent algorithms for lasso penalized regression},} \textit{The Annals of Applied Statistics}, 2(1), 224--244.

\bibitem[{Yang and Zou(2013)}]{YangZou2013}
Yang, Y. and Zou, H. (2013), \enquote{{An efficient algorithm for computing the HHSVM and its generalizations},} \textit{Journal of Computational and Graphical Statistics}, 22(2), 396--415.


\bibitem[{Zhang(2004)}]{Zhang2004}
Zhang, T. (2004), \enquote{{Statistical behavior and consistency of classification methods based on convex risk minimization},} \textit{The Annals of Statistics}, 32(1), 56--134.

\bibitem[{Zhou and Lange(2010)}]{ZhouLange2010}
Zhou, H. and Lange, K. (2010), \enquote{{MM algorithms for some discrete multivariate distributions},} \textit{Journal of Computational and Graphical Statistics}, 19(3), 645--665.

\bibitem[{Zhu and Hasite(2005)}]{ZhuHastie2005}
Zhu, J. and Hastie, T. (2005), \enquote{{Kernel logistic regression and the import vector machine},} \textit{Journal of Computational and Graphical Statistics}, 14(1), 185--205.

\bibitem[{Zou and Li(2008)}]{ZouLi2008}
Zou, H. and Li, R. (2008), \enquote{{One-step sparse estimates in nonconcave penalized likelihood models},} \textit{The Annals of Statistics}, 36(4), 1509-1533.
\end{thebibliography}
\end{document}